\documentclass[12pt]{article}

\usepackage{lmodern}

\usepackage{amsmath,amssymb}

\usepackage{booktabs}
\usepackage{longtable}
\usepackage{array}
\usepackage{calc}
\newcommand{\real}[1]{#1}

\usepackage{graphicx}
\usepackage{float}
\graphicspath{{figures/}}

\usepackage{xcolor}

\usepackage[hidelinks]{hyperref}

\usepackage[numbers,sort&compress]{natbib}

\usepackage{microtype}
\usepackage[margin=2.5cm]{geometry}

\begin{document}

\title{Direct Clinical Joint Angle Extraction from Parametric Body Model Rotation Matrices}
\author{J. Kardolus$^1$ \and D. Hendriks$^1$ \and J. Jansen$^2$}
\date{
  $^1$Babon Innovations B.V., Utrecht, The Netherlands \\
  $^2$Research Group Lifestyle and Health, HU University of Applied Sciences Utrecht, The Netherlands
}
\maketitle

\begin{abstract}
Quantitative joint angles are rarely available in routine care because the tools are slow, costly, or confined to a laboratory. We show that clinical joint angles can be read directly from the per-segment rotation matrices a parametric body model already produces, with no inverse-kinematics or musculoskeletal-model fitting step. On the OpenCap LabValidation cohort, using the GEM-X body-model estimator on single-smartphone video, our pooled mean absolute error is 4.50° over the fifteen joint angles that match the OpenCap Monocular reference set, the same accuracy range as OpenCap Monocular's 4.8° on the same cohort and reference standard, from a much simpler pipeline. The step that connects a body model to clinical angles is a small calibration table rather than an optimisation, so the same procedure transfers unchanged to other body models: repeating it on SAM 3D Body, changing only the table, gives 4.66°, statistically indistinguishable from GEM-X, and runs in real time from a live single-camera stream. The method needs no per-recording inputs beyond the video itself: no participant height, no camera-intrinsics database, no per-subject model scaling. This broadens where movement analysis is practical, from in-clinic and at-home recording to telerehabilitation and large-scale decentralised studies.
\end{abstract}

\section{Introduction}

Three decades of growth in computing power \cite{schaller1997moore} have established artificial intelligence as one of the fastest-growing technical fields \cite{jordan2015ml}, and deep neural networks \cite{krizhevsky2012alexnet} have since become the dominant approach to machine perception \cite{lecun2015deeplearning}. Much of the investment that followed was directed toward robotics and autonomous systems, among the largest application areas drawing AI funding \cite{aiindex2025}, and toward computer vision, whose methods apply across an unusually broad range of tasks \cite{voulodimos2018cvreview}. Within vision, human pose estimation emerged as the problem of recovering a person's posture and movement from ordinary images \cite{zheng2023hpesurvey}: a capability robots need to perceive the people around them, with relevance well beyond robotics, in fields such as medicine and sport science.

Both have long sought to quantify human movement as they quantify other markers of health. Instrumented gait analysis is an established clinical discipline \cite{baker2006gait}, but it depends on marker-based optical motion capture: equipment costing on the order of a hundred thousand euros, a dedicated laboratory, trained operators, and cohorts of a few dozen participants. Quantitative joint kinematics have therefore remained largely absent from routine care, field studies, and everyday sport and rehabilitation, motivating a search for markerless alternatives.

OpenCap was among the larger open-source efforts in this direction, recovering movement from synchronised smartphone video and open musculoskeletal modelling \cite{uhlrich2023opencap, seth2018opensim}, yet it still requires two or more cameras, a per-session calibration, and non-trivial analysis software. Such constraints point toward monocular, single-camera analysis: if joint kinematics could be recovered from one ordinary video, almost any recording could become movement data.

Monocular accuracy has been the limiting factor. A recent benchmark evaluated eleven monocular pose estimators, each over a 2.2-million-frame dataset against marker-based ground truth, and found that none met the five-degree joint-angle error taken as the threshold for correct clinical interpretation; its authors recommended estimators built around an integrated kinematic body model as a prior on plausible human motion \cite{rode2025physio2m}. Parametric body models supply that prior, representing the body as a connected chain of rigid segments \cite{loper2015smpl, pavlakos2019smplx, park2025atlas, ferguson2025mhr, bregier2025anny, saito2026soma} from which estimators recover a per-segment rotation matrix from a single video \cite{kanazawa2018hmr, shin2024wham, li2025genmo, yang2026sam3d}. Much of this work now originates in large industrial AI laboratories \cite{li2025genmo, yang2026sam3d}, and it is that robotics- and vision-driven research this paper connects to clinical measurement.

A rotation matrix, however, is not yet a clinical angle. Several monocular pipelines turn body-model output into clinical joint angles by adding a musculoskeletal-model fitting stage: OpenCap's monocular extension runs OpenSim inverse kinematics on virtual markers \cite{gilon2026opencapmono}, the Portable Biomechanics Laboratory fits a differentiable MuJoCo model to keypoints \cite{peiffer2025pbl}, and SKEL re-rigs a body model with a biomechanical skeleton \cite{keller2023skel, werling2023addbio}, while BioPose and HSMR regress biomechanically constrained poses directly \cite{koleini2025biopose, xia2025hsmr}. We observe that this stage is not always necessary: a parametric body model already specifies the relative orientation of every parent--child segment pair, and the International Society of Biomechanics conventions \cite{grood1983jcs, wu2002isb1, wu2005isb2} describe the same orientation in another reference frame, so moving between them is a coordinate transformation, not an optimisation. The swing-twist decomposition itself is not new: HybrIK uses it inside a monocular pose estimator, deriving the swing from predicted joint positions and regressing the twist to reconstruct a body mesh \cite{li2021hybrik}. We apply the same decomposition at the opposite end of the pipeline, reading clinical angles out of the rotations a body model already produced, with no reconstruction or inverse-kinematics step.

This paper contributes not a new pipeline but a method for that transformation, turning the chained per-segment rotation matrices a body model already produces directly into standardised clinical joint angles. What is new is the combination of three elements: a single per-joint extraction procedure with no joint-type-specific code; a compact calibration table, about ten numbers per joint, linking one body model to the clinical convention; and a demonstration that this is sufficient, on a standard dataset against the same reference standard as established methods, to obtain clinical angles from body-model rotations alone. We validate on a monocular instantiation, GEM-X with the SOMA skeleton on OpenCap LabValidation \cite{li2025genmo, saito2026soma, uhlrich2023opencap}, and compare against OpenCap Monocular \cite{gilon2026opencapmono} on its own cohort and reference standard. To test whether the procedure transfers across body models, we repeat it on SAM 3D Body \cite{yang2026sam3d}, changing only the calibration table; because that estimator runs in real time, the same clinical angles are available from a live single-camera stream. Other modalities, including multi-camera, inertial, and depth-sensor inputs, are compatible with the same procedure, and their validation is left to future work.

\section{Methods}
\subsection{Pipeline Overview}

The pipeline has two stages. The first turns sensor data into per-segment rotation matrices via a parametric body model. The second turns those rotation matrices into clinical joint angles in the International Society of Biomechanics convention. This paper is about the second stage. The first stage is treated as a supplier of inputs to the second; any body-model estimator that produces, from sensor data, per-segment rotation matrices in a kinematic chain, together with the body model's rest pose, can serve in this role.

We validate on two body-model estimators. The primary instantiation, used to describe the method below, is GEM-X \cite{li2025genmo} with the SOMA skeleton convention \cite{saito2026soma} and monocular video as the sensor input; the same procedure is repeated on SAM 3D Body \cite{yang2026sam3d} with the MHR skeleton as a cross-body-model check (Appendix C). SOMA defines a kinematic chain of 77 joints connecting 78 rigid segments. For each input frame, the estimator emits a 3×3 rotation matrix per segment together with a set of rotation matrices describing the body model's rest pose. These rotations, the rest pose, and the kinematic chain are the entire input to the extraction stage described in the remainder of this section.

\subsection{Clinical Angle Extraction}

The extraction turns one frame of body-model output into one clinical joint angle in three steps: isolate the joint's motion, decompose it along three calibrated axes into three magnitudes, and combine those magnitudes into the clinical angle. Every step-specific value (three axes, three weights, a left/right sign, an offset) comes from a small static \emph{calibration table}, produced once per body model by the procedure in Section 2.3.

Each joint is a parent-child segment pair in the body model's kinematic chain, following the conventional anatomical pairings (Appendix D). The pelvis is the root and has no parent.

\textbf{Step 1: isolate the joint motion.} Each frame, the estimator gives a 3×3 world-orientation matrix per segment. For a joint we want only the rotation between its two segments, and only the part caused by movement: the rest pose is non-zero, so these matrices are not identity even when the subject stands still. We express the child in the parent's frame and subtract the same quantity at the rest pose:

\[R_{\text{motion}} \;=\; R_{\text{local}} \, R_{\text{rest,local}}^{-1}, \quad \text{where} \quad R_{\text{local}} \;=\; R_{\text{parent}}^T R_{\text{child}}.\]

$R_{\text{local}}$ is the child's orientation in the parent's frame; $R_{\text{rest,local}}$ is that same local rotation at the rest pose, fixed per joint per body model. Subtracting it makes $R_{\text{motion}}$ the identity in the rest pose, whatever rest configuration the body model uses.

The pelvis, having no parent, gets the rest-pose correction on its global rotation directly; we remove its horizontal heading so tilt and obliquity depend only on intrinsic orientation, and take pelvis transverse rotation from the centre-of-mass travel direction (undefined below a minimum walking speed).

\textbf{Step 2: decompose into swing and twist.} The body model's axes carry no anatomical meaning, so flexion cannot be read off the matrix directly. The table supplies three unit axes: a \emph{twist axis} $e_1$ (the bone's long axis) and two axes $e_2, e_3$ spanning the perpendicular \emph{swing plane}. We split $R_{\text{motion}}$ into the swing that carries $e_1$ to its new direction and the residual twist about $e_1$, giving three magnitudes $(c_0, c_1, c_2)$:

\[c_0 \;=\; \text{twist about } e_1, \qquad c_1 \;=\; \boldsymbol{\phi}_{\text{swing}}\!\cdot e_2, \qquad c_2 \;=\; \boldsymbol{\phi}_{\text{swing}}\!\cdot e_3,\]

where $\boldsymbol{\phi}_{\text{swing}}$ is the swing rotation vector (direction = swing axis, length = swing angle). Unlike a fixed three-axis Euler decomposition, whose cross-axis coupling grows with amplitude, this keeps a joint's dominant rotation on one component across a wide range of motion: a ball joint's swing does not leak into the twist channel as the pose changes. The decomposition is rotation-only and closed-form.

\textbf{Step 3: produce the clinical joint angle.} The three magnitudes are not yet the clinical angle. The rest pose does not match the clinical neutral (the SOMA T-pose holds the arms out sideways, so shoulder offsets are large; the knee rest pose nearly matches neutral, so its offset is small), and each magnitude's amplitude need not match the reference angle's. One linear step applies both corrections:

\[\theta_{\text{clinical}} \;=\; w_1 \, c_0 \;+\; w_2 \, c_1 \;+\; w_3 \, c_2 \;+\; \delta.\]

The weights $(w_1, w_2, w_3)$ scale the magnitudes (typically one near $\pm 1$, the rest near zero) and $\delta$ is the rest-pose offset; both come from the table. For bilateral joints a per-side sign is multiplied into the magnitudes, so one set of weights and offset serves both sides.

The same three steps run on every joint, with no joint-specific code. Joints with fewer than three clinical directions (knee and elbow are single-axis, ankle has two) emit the unused magnitudes as diagnostic channels, not clinical outputs.

The output angle series is low-pass filtered at 6 Hz, the conventional biomechanics cut-off \cite{winter2009}. No virtual markers, surface meshes, or musculoskeletal models are built at any stage.

\subsection{Calibration Procedure}

The calibration table is produced once per body model. We compare the extraction output of Section 2.2 against a reference dataset (typically marker-based motion capture processed through inverse kinematics in a musculoskeletal modelling package such as OpenSim) and determine the table values in three offline phases.

\textbf{Phase 1: identify the axes.} For each joint we fix the twist axis $e_1$ to the bone's long axis and evaluate candidate swing-plane axes together with the channel assignment (which swing-twist component corresponds to flexion, abduction, and rotation) and per-side sign, selecting the combination with the highest correlation to the reference. The initial offset is the average angular difference between the decoded angle and the reference, averaged across the left and right sides for symmetry.

\textbf{Phase 2: refine the axes.} A body model's internal axes do not perfectly coincide with anatomical rotation axes, and the residual misalignment shows up as cross-channel coupling (knee flexion contaminating the adduction channel, for example). Phase 2 keeps Phase 1's choices and lets each axis tilt by a small amount. We evaluate small candidate adjustments to the three axes and select the adjustment that gives the closest agreement with the reference, both in shape and in amplitude. Bilateral joints share a single set of refined axes applied symmetrically to both sides.

\textbf{Phase 3: determine the weights and offset.} Phase 3 produces the two corrections described in Step 3: the rest-pose offset $\delta$ and the three amplitude weights $(w_1, w_2, w_3)$. We fit both together per clinical joint angle by least squares against the reference, with the offset a free intercept for every joint --- the same rule everywhere, no joint-specific offset handling. The weight fit carries a ridge penalty on the three weights (not the intercept): a single-axis joint's swing-twist components are near-collinear, so an unpenalised fit can choose large cancelling weights that track well in-range but amplify small extrapolations at the edges of the range into large errors (a deep squat reading far past anatomical knee flexion, for example). The penalty bounds the weight magnitudes with no measurable in-range accuracy cost, so extreme-range motion extrapolates smoothly. Bilateral joints share a single set of weights and offset; a per-side sign is multiplied into the input.

\subsection{Validation Protocol}

We validate on OpenCap LabValidation \cite{uhlrich2023opencap}: nine subjects, 720 multi-camera trials of walking, squats, sit-to-stand, and drop jumps, recorded simultaneously with marker-based motion capture and synchronised smartphone video. Reference joint angles come from the marker-based capture through OpenSim inverse kinematics \cite{seth2018opensim} on the LaiUhlrich2022 musculoskeletal model \cite{uhlrich2023opencap, lai2017antagonist} at 100 Hz. The method under evaluation sees only the monocular smartphone video, one camera at a time: the body-model estimator turns it into per-segment rotation matrices, which feed the Section 2.2 extraction. The external comparison target is OpenCap Monocular \cite{gilon2026opencapmono}.

\textbf{Camera input.} The dataset's video comes from five synchronised iPhone 12 Pro cameras, recording at 720×1280 and 60 Hz, placed 1.5 m off the ground and 3 m from the capture volume, one at 0° and pairs at ±45° and ±70° (five positions in total) \cite{uhlrich2023opencap}; the marker data (an eight-camera 100 Hz optical system, 31 markers \cite{uhlrich2023opencap}) is never a pipeline input, only the reference. For the matched comparison in Section 3.1 we use the single 45° view (Cam1), the same iPhone camera model and 45° viewing angle OpenCap Monocular validated on, so the two methods meet on the same cohort, camera, viewing angle, and reference standard. The per-view results in Appendix B (Table B2) instead pool all five views, each used as a separate monocular input, which adds held-out data per subject and gives an accuracy estimate that does not depend on one fixed camera placement.

Two points govern how residuals are computed. Video and reference are not synchronised at the millisecond level, so before comparing them we apply a per-trial sub-frame alignment that minimises the centred residual pooled across bilateral knee, hip, and ankle flexion (the alignment shifts the reference, not the prediction). For each clinical DOF we report Pearson r, RMSE, and MAE; we also report centred MAE (cMAE, per-trial bias removed before averaging) to separate within-trial waveform tracking from constant offset. Correlation is noise-dominated on channels with a small reference range, so the absolute-error metrics are the reliable accuracy indicator there.

All accuracy figures are held-out under leave-one-subject-out (LOSO) cross-validation: nine folds, the complete three-phase calibration of Section 2.3 fit on the eight training subjects of each fold, the held-out subject's trials scored against it. This mirrors production, where every new patient is a held-out subject; deployed accuracy is, if anything, slightly better, since deployment uses all nine subjects rather than the eight available in any single fold. No subjects, trials, or DOFs are excluded; outlier behaviour surfaces in the standard-deviation column.

\section{Results}
The detail sits in the appendices: the full per-DOF tables, the channel-by-channel breakdown, and the cycle-aligned waveforms for the main GEM-X validation in Appendix B, the calibration table in Appendix A, and the cross-body-model comparison in Appendix C. This section reports the headline.

\subsection{Matched-condition comparison with the OpenCap Monocular reference}

OpenCap Monocular validates on the same cohort and reference standard used here, which makes it the most direct external comparison for this work; it reports a pooled mean absolute error (MAE) of 4.8° over 18 joint angles from a single 45° smartphone view on non-jumping activities \cite{gilon2026opencapmono}. We match those conditions, comparing on the fifteen of their joint angles that have a direct counterpart in our set (Table B1), and obtain a pooled MAE of 4.50°, or 3.11° once each trial's constant offset is removed. A bootstrap over the nine subjects (20,000 draws) gives a 95\% confidence interval of 4.28°--4.78°, which sits at or just below the 4.8° reference, so the two methods are in the same accuracy range on this cohort. OpenCap reports no centred MAE.

The two pipelines reach this accuracy from different inputs. OpenCap Monocular adds five processing stages to monocular pose estimation: a pose-refinement optimisation, an activity classifier, per-subject scaling of an OpenSim model, a camera-intrinsics database, and a participant-height query. The extraction reported here uses only the per-segment rotation matrices, with no per-recording inputs.

\begin{figure}[H]
  \centering
  \includegraphics[width=\textwidth]{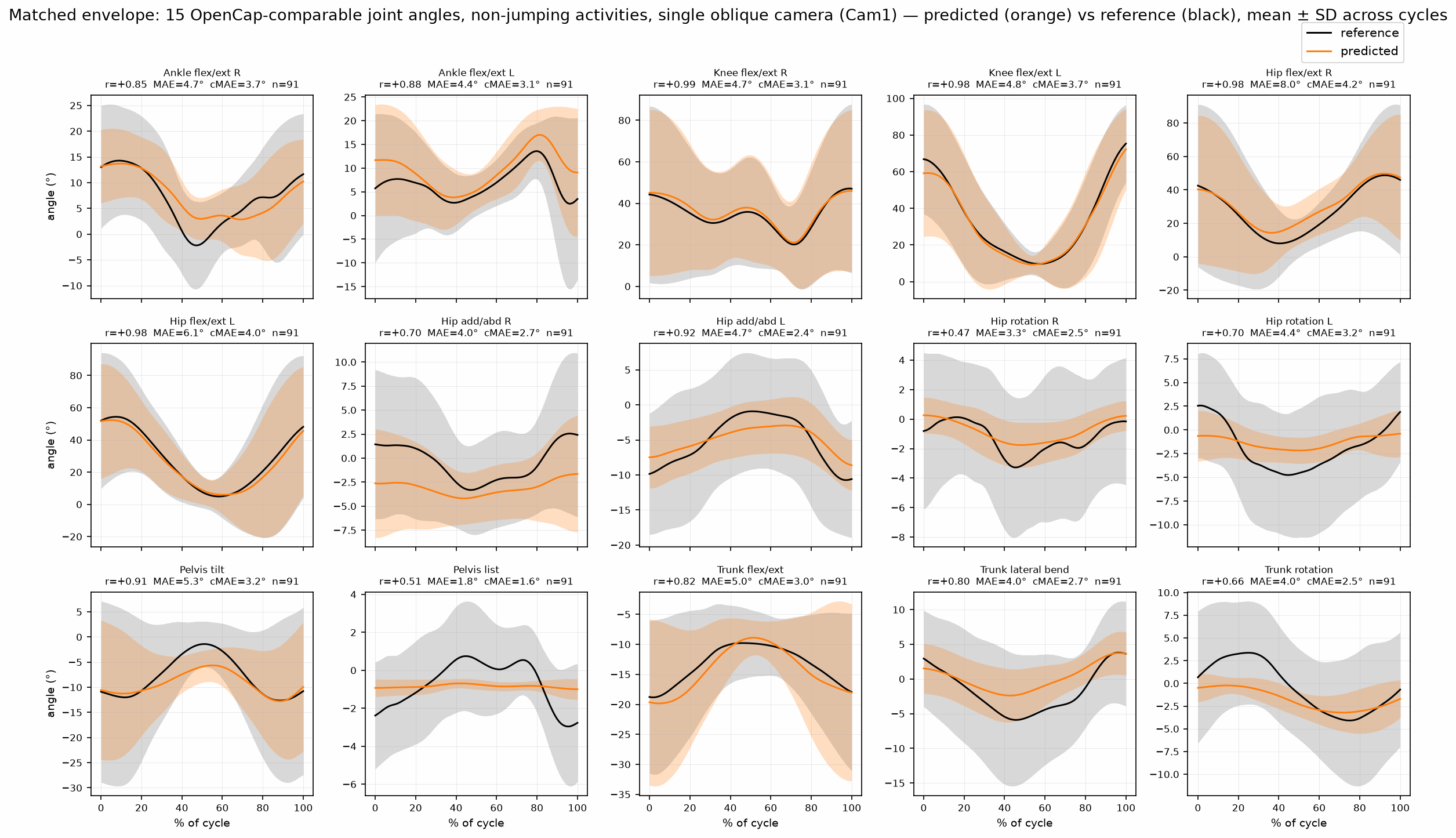}
  \caption{The matched envelope: the fifteen OpenCap-comparable joint angles behind the headline comparison, predicted (orange) over the OpenSim reference (black), mean ± 1 SD across cycles, pooled over the non-jumping activities (walking, squats, sit-to-stand) at the single 45° camera (Cam1). This is the exact fifteen-DOF, non-jumping, single-camera envelope that yields the 4.50° pooled MAE. The flexion channels (knee, hip, ankle) track closely; the wider bands on the out-of-plane rotation channels (hip and pelvis rotation, trunk) are the monocular depth-ambiguity limit discussed in Section 4.3. Each panel's r, MAE, and centred MAE are pooled cycle-level values under the full-cohort calibration; the authoritative per-DOF held-out (LOSO) accuracy is in Table B1.}
  \label{fig:matched_envelope}
\end{figure}

\subsection{Accuracy regimes}

Three regimes emerge from Table B1. The \textbf{strong} regime (held-out correlation roughly 0.80 or higher) covers bilateral knee flexion, bilateral hip flexion, bilateral shoulder flexion, bilateral ankle dorsiflexion, and elbow flexion. The \textbf{moderate} regime (correlation around 0.60 to 0.80) covers hip adduction and trunk lateral bend. The \textbf{limited} regime (correlation below 0.55) covers internal and external rotation at the hip and shoulder, shoulder abduction, ankle inversion and eversion, and pelvis tilt and list; these are the channels where monocular depth ambiguity dominates the underlying signal.

\subsection{Cross-body-model check}

To probe the body-model-agnostic claim we repeated the identical procedure on a second estimator and skeleton, SAM 3D Body with the MHR body model, on the same cohort and reference standard, changing only the calibration table. On the matched fifteen-DOF envelope its mean absolute error is 4.66° against GEM-X's 4.50°; paired by subject the difference is +0.20° (95\% CI −0.05° to +0.48°), an interval that spans zero, so the two skeletons are statistically indistinguishable on this cohort. Because SAM 3D Body runs in real time, the same clinical angles are available from a live single-camera stream rather than offline batch. The full per-activity breakdown and cycle-aligned waveforms are in Appendix C.

\section{Discussion}
\subsection{Implications of direct extraction}

The results indicate that an additional musculoskeletal-model fitting stage is not strictly necessary in a body-model-based pipeline if the goal is to produce clinical joint angles in established conventions. A parametric body model already encodes per-segment orientations in a defined kinematic chain, and the mapping to a clinical angle convention can be expressed as a calibration problem rather than an optimisation. The practical consequences of this observation are concrete:

\begin{itemize}
\item
  \textbf{No biomechanics-specific dependencies at runtime.} This is the largest practical consequence. The extraction is a small set of matrix multiplications and a few scalar operations per joint, expressible in any language with basic linear algebra. It does not call into a biomechanics library, an inverse-kinematics package, an optimiser, or a numerical solver. Once the per-segment rotation matrices have been produced by the body-model estimator, every downstream step is plain arithmetic. The same extraction code runs unchanged on a smartphone, an in-clinic workstation, a server, an embedded device, or a browser, with no biomechanics software to install. In a body-model-based pipeline, the extraction step is no longer a bottleneck.
\item
  \textbf{Runtime cost.} The extraction adds about one millisecond per frame on top of whatever the upstream body-model estimator costs. A musculoskeletal-model fitting stage typically adds seconds to minutes per recording, depending on the optimiser and the model.
\item
  \textbf{Determinism.} Identical inputs produce identical outputs; the extraction is closed-form and non-iterative.
\item
  \textbf{Calibration cost.} All three calibration phases run once per body model, offline. The result is a static table that is then applied to every subsequent recording, with no per-recording calibration step.
\item
  \textbf{Body-model agnosticism.} Any pose estimator that produces per-segment rotation matrices can supply inputs to the extraction; only the calibration table needs to be regenerated if the body model changes. Empirical work in the body-model literature \cite{austin2026rigs} suggests that swapping the underlying skeleton (for example, from SMPL to MHR) can improve reconstruction quality more than scaling the pose-estimation network, and the extraction described here inherits any such gains automatically.
\item
  \textbf{Implementation simplicity.} The same code path runs on every joint; there is no joint-type-specific branching in the runtime code. And because the method takes no per-recording inputs --- no participant height, no camera intrinsics, no per-subject model scaling, no per-camera tuning --- the same fixed calibration table runs on any monocular video regardless of the camera or viewing angle, so it can be pointed at essentially any video source.
\end{itemize}

\subsection{What the calibration table reveals}

The calibration table (Appendix A) is itself informative.

Which swing-twist component carries each clinical angle is not predictable from the body model's own axis labels. The twist axis is anchored to the bone long axis, but which of the two swing-plane projections (or the twist itself) maps to flexion, abduction, or rotation differs from joint to joint (Table A1), and the swing-plane orientation that aligns cleanly with the clinical convention (Table A2) must be found from data because the body-model's internal axes carry no anatomical meaning. The projection weights (Table A3) are correspondingly informative: after ridge regularisation each clinical angle is carried almost entirely by a single swing-twist component with a weight near ±1, and the small off-component weights (mostly below 0.4) are the residual cross-axis coupling the calibration absorbs rather than the dominant signal.

Sign corrections are not symmetric between the left and right sides for every channel. Hip rotation takes opposite signs on the two sides, while trunk and pelvis channels share a single sign. This reflects mirrored mediolateral axis conventions in the body model and is resolved by the calibration empirically, rather than by anatomical assumption.

\subsection{Limitations}

\textbf{Single dataset.} The validation uses the OpenCap LabValidation cohort: nine healthy young adults in a controlled lab setting. Accuracy on further body models (SMPL, SMPL-X, and others) and on pathological movement patterns remains to be characterised in future work.

\textbf{Out-of-plane rotation channels.} Internal and external rotation correlations are lower than flexion correlations (roughly 0.25 to 0.49 against 0.79 to 0.99). The limiting factor is the depth ambiguity inherent in monocular pose estimation and is not specific to the extraction procedure described here; multi-camera or depth-sensor instantiations of the same extraction would be expected to recover these channels better.

\textbf{Calibration range coverage.} A calibrated zero-point is only as well-determined as the range the reference dataset exercises for that joint. The OpenCap LabValidation activities (walking, squats, sit-to-stand, drop jumps) are lower-limb dominant, so the arm rarely reaches full extension: the elbow flexion channel is tracked well within its observed range but has few near-straight frames to pin its zero, so its absolute offset is less certain than a lower-limb channel calibrated over the same protocol. A reference set with richer upper-limb range of motion would tighten these zero-points; the tracking (shape) is unaffected.

\subsection{Camera sampling and fast movements}

Accuracy is lowest on the fastest activity, the drop jump (Appendix B, Table B2; Appendix C), and part of that gap is a property of the camera rather than of the extraction. The LabValidation video is recorded at 60 Hz on a consumer phone. During a jump landing a limb moves far between successive frames and can blur within a single frame, so the body-model estimator receives fewer and softer samples of the pose at exactly the moments when the joint angles change fastest. Both effects, temporal undersampling \cite{dunn2023vfi} and motion blur \cite{ma2024posebench}, degrade the pose input before the extraction runs, so they cap accuracy on rapid movement independently of the method. A user who needs fast-movement accuracy can address this at capture time with a higher frame rate and a shorter shutter (exposure) time, both of which modern phone cameras support; the extraction itself is unchanged. This is a point about the shared input, not about the comparison: it does not affect the matched OpenCap Monocular comparison (Section 3.1), which uses the same 60 Hz iPhone footage for both systems and is restricted to non-jumping activities.

\subsection{Related monocular methods}

Several other monocular methods give useful context, though a direct numeric comparison is constrained by differences in datasets, metrics, and degree-of-freedom selection. The Portable Biomechanics Laboratory \cite{peiffer2025pbl} fits a differentiable MuJoCo biomechanical model to video keypoints and validates clinically across multiple impaired populations, an end-to-end alternative to the body-model route taken here. SKEL \cite{keller2023skel} re-rigs SMPL with a biomechanical skeleton via AddBiomechanics \cite{werling2023addbio}, the kind of body model the extraction can target directly. Physio2.2M \cite{rode2025physio2m} benchmarks eleven open-source markerless estimators across 2.2 million RGB frames, a population-scale reference for estimator quality.

\subsection{Relationships to adjacent work}

GaitDynamics \cite{tan2026gaitdynamics} is a recent method that predicts ground reaction forces from joint angles expressed in a standard biomechanical convention, using a transformer model. The output of the extraction described in this paper is in the same convention, which means a body-model-to-forces chain can in principle be constructed by composing the two methods directly, without an explicit inverse-kinematics step in between.

A further instance is GaitEncoder \cite{magruder2026gaitencoder}, a variational-autoencoder foundation model that compresses whole-body walking kinematics into a compact latent representation and a scalar impairment score for a range of downstream clinical tasks. Like GaitDynamics, it consumes joint angles in a standard clinical convention. The extraction described here produces angles in that same convention directly from body-model rotations, without the musculoskeletal-model fitting stage, so its output can feed such a model through simple resampling of the angle series into gait cycles, with no return to inverse kinematics.

The extraction's input is per-segment rotation matrices from any body model. Generative motion models, including text-conditioned models such as Kimodo \cite{rempe2026kimodo}, produce rotation matrices in formats compatible with the extraction, opening a path to applying the same clinical-angle interpretation to synthetic or generated movement.

Unification layers extend the reach further. The SOMA convention itself \cite{saito2026soma} is designed to recover a unified set of skeleton rotations from posed vertices of multiple underlying body models (SMPL, SMPL-X, ATLAS, MHR, and Anny). An estimator producing a posed mesh in any of these conventions can therefore feed the extraction through the SOMA abstraction, with only the per-body-model calibration table needing to be regenerated if the underlying skeleton definition differs.

Clinical work in this space has historically used spatiotemporal gait metrics rather than full joint kinematics, in part because robust joint-angle extraction from accessible sensors was difficult. For example, Banarjee et al. \cite{banarjee2026fall} apply world-aware human mesh recovery to the Timed Up and Go test and report on spatiotemporal parameters; the extraction here adds a route to clinical joint angles from the same input, enabling musculoskeletal-level analyses in the same setting.

\section{Conclusion}

Clinical joint angles can be obtained directly from the rotation matrices that a parametric body model already produces, with no inverse-kinematics or musculoskeletal-model fitting stage. The values that do this live in a static calibration table of about ten numbers per joint, produced once per body model from a reference dataset. The extraction itself is deterministic, computationally negligible at runtime, and runs unchanged on anything from a smartphone to a server.

Validated on a monocular instantiation against the same reference dataset used by OpenCap Monocular, the pooled mean absolute error across the fifteen rotational joint angles we measure as anatomically comparable to the OpenCap Monocular reference set is 4.50°, the same accuracy range as OpenCap Monocular's 4.8° on that dataset, from a simpler pipeline. Repeating the same procedure on a second body model, SAM 3D Body, changing only the calibration table, gives 4.66°, statistically indistinguishable from the first: evidence that the extraction is not tied to one body model. Because that estimator runs in real time, the same clinical angles are available from a live single-camera stream rather than offline batch.

Quantitative joint kinematics have long been rare in routine clinical care because measuring them required a specialised laboratory, expensive equipment, and a multi-stage software pipeline. Markerless methods such as OpenCap already removed the laboratory and the expensive equipment, replacing them with ordinary video; the work presented here removes the piece that remained, the multi-stage musculoskeletal-fitting software, leaving only the body-model estimator between a video and a clinical joint angle. The same extraction works with any body-model estimator, so a newer or faster one can be swapped in without changing the method. We hope this lowers the barrier far enough that biomechanical analysis can be a routine assessment in clinics, in homes, in sports and rehabilitation programmes, and in the many adjacent fields where movement carries diagnostic information but the means to measure it precisely have so far been out of reach.

\bibliographystyle{unsrtnat}
\bibliography{refs}

\appendix
\section{Full calibration tables}

The calibration procedure of Section 2.3 produces three sets of values per body model: the component-to-clinical-angle assignment and per-side sign (Phase 1), the three swing-twist axes (Phase 2), and the three projection weights and offset per clinical joint angle (Phase 3). Tables A1--A3 list all three sets for the SOMA skeleton convention with the GEM-X body-model estimator. The swing-twist decomposition replaces the classical sequential Cardan axis sequence with a fixed per-joint component order: each joint carries one twist axis $e_1$ (its bone long axis) and two swing axes $e_2, e_3$ in the plane perpendicular to it, and the decomposition returns (twist, swing·$e_2$, swing·$e_3$) in that fixed order (Section 2.2).

\begin{longtable}{@{}llrcrr@{}}
\caption{\textbf{Table A1.} Phase 1 selections per joint. For each clinical joint angle the table lists the index of the swing-twist component identified as the dominant input (idx: 0 = twist, 1 = swing·$e_2$, 2 = swing·$e_3$), the per-side sign correction (mirroring left and right for bilateral joints), the offset $\delta$, and the in-sample Pearson correlation against the reference (r). Channels marked ``info'' are diagnostic outputs of single-axis and two-axis joints (knee, elbow, ankle) and are not used as clinical outputs. Pelvis transverse rotation is reported from the centre-of-mass-yaw surrogate (Section 2.2).}
\label{tab:a1}\\
\toprule\noalign{}
Joint & DOF & idx & sign & offset (°) & r \\
\midrule\noalign{}
\endfirsthead
\toprule\noalign{}
Joint & DOF & idx & sign & offset (°) & r \\
\midrule\noalign{}
\endhead
\bottomrule\noalign{}
\endlastfoot
hip\_L & adduction & 0 & -1 & −1.397 & 0.547 \\
hip\_L & flexion & 1 & -1 & +11.752 & 0.923 \\
hip\_L & rotation & 2 & -1 & +1.503 & 0.515 \\
hip\_R & adduction & 0 & +1 & −1.397 & 0.428 \\
hip\_R & flexion & 1 & -1 & +11.752 & 0.931 \\
hip\_R & rotation & 2 & +1 & +1.503 & 0.367 \\
shoulder\_L & adduction & 0 & -1 & −50.288 & 0.431 \\
shoulder\_L & flexion & 1 & -1 & +34.220 & 0.812 \\
shoulder\_L & rotation & 2 & -1 & +21.581 & 0.466 \\
shoulder\_R & adduction & 0 & +1 & −50.288 & 0.451 \\
shoulder\_R & flexion & 1 & -1 & +34.220 & 0.799 \\
shoulder\_R & rotation & 2 & +1 & +21.581 & 0.467 \\
trunk & adduction & 0 & +1 & +1.449 & 0.499 \\
trunk & flexion & 1 & -1 & −6.645 & 0.679 \\
trunk & rotation & 2 & +1 & −1.575 & 0.443 \\
pelvis & flexion & 0 & -1 & −6.094 & 0.561 \\
pelvis & transverse (COM-yaw surrogate) & 1 & -1 & −0.402 & info \\
pelvis & adduction & 2 & +1 & −0.883 & 0.397 \\
knee\_L & flexion & 0 & +1 & +2.902 & 0.976 \\
knee\_L & off-axis (informational) & 1 & +1 & +0.000 & info \\
knee\_L & off-axis (informational) & 2 & +1 & +0.000 & info \\
knee\_R & flexion & 0 & +1 & +2.902 & 0.981 \\
knee\_R & off-axis (informational) & 1 & +1 & +0.000 & info \\
knee\_R & off-axis (informational) & 2 & +1 & +0.000 & info \\
elbow\_L & off-axis (informational) & 0 & +1 & +0.000 & info \\
elbow\_L & off-axis (informational) & 1 & +1 & +0.000 & info \\
elbow\_L & flexion & 2 & +1 & +24.561 & 0.717 \\
elbow\_R & off-axis (informational) & 0 & +1 & +0.000 & info \\
elbow\_R & off-axis (informational) & 1 & +1 & +0.000 & info \\
elbow\_R & flexion & 2 & +1 & +24.561 & 0.716 \\
ankle\_L & off-axis (informational) & 0 & +1 & +0.000 & info \\
ankle\_L & flexion & 1 & -1 & +6.698 & 0.837 \\
ankle\_L & adduction & 2 & +1 & −5.369 & 0.243 \\
ankle\_R & off-axis (informational) & 0 & +1 & +0.000 & info \\
ankle\_R & flexion & 1 & -1 & +6.698 & 0.867 \\
ankle\_R & adduction & 2 & +1 & −5.369 & 0.425 \\
\end{longtable}

\begin{longtable}{@{}
  >{\raggedright\arraybackslash}p{(\columnwidth - 6\tabcolsep) * \real{0.4375}}
  >{\raggedright\arraybackslash}p{(\columnwidth - 6\tabcolsep) * \real{0.1875}}
  >{\raggedright\arraybackslash}p{(\columnwidth - 6\tabcolsep) * \real{0.1875}}
  >{\raggedright\arraybackslash}p{(\columnwidth - 6\tabcolsep) * \real{0.1875}}@{}}
\caption{\textbf{Table A2.} Phase 2 swing-twist axes per joint: $e_1$ is the twist axis (the bone long axis), $e_2$ and $e_3$ span the perpendicular swing plane. Each is a unit vector in the body model's local segment frame. Bilateral joint pairs (hip, shoulder, knee, elbow, ankle) share a single set of axes applied symmetrically to both sides.}
\label{tab:a2}\\
\toprule\noalign{}
\begin{minipage}[b]{\linewidth}\raggedright
Joint
\end{minipage} & \begin{minipage}[b]{\linewidth}\raggedright
Twist axis $e_1$
\end{minipage} & \begin{minipage}[b]{\linewidth}\raggedright
Swing axis $e_2$
\end{minipage} & \begin{minipage}[b]{\linewidth}\raggedright
Swing axis $e_3$
\end{minipage} \\
\midrule\noalign{}
\endfirsthead
\toprule\noalign{}
\begin{minipage}[b]{\linewidth}\raggedright
Joint
\end{minipage} & \begin{minipage}[b]{\linewidth}\raggedright
Twist axis $e_1$
\end{minipage} & \begin{minipage}[b]{\linewidth}\raggedright
Swing axis $e_2$
\end{minipage} & \begin{minipage}[b]{\linewidth}\raggedright
Swing axis $e_3$
\end{minipage} \\
\midrule\noalign{}
\endhead
\bottomrule\noalign{}
\endlastfoot
hip (L+R) & (+0.145, +0.989, −0.026) & (+0.009, +0.025, +1.000) & (+0.723, +0.691, +0.004) \\
shoulder (L+R) & (+0.514, +0.858, −0.014) & (−0.007, +0.020, +1.000) & (+0.999, +0.007, −0.036) \\
trunk & (−0.240, +0.971, +0.015) & (−0.198, +0.693, +0.693) & (+0.970, +0.243, +0.029) \\
pelvis & (+0.991, −0.080, +0.109) & (+0.700, +0.700, +0.141) & (+0.577, +0.577, +0.577) \\
knee (L+R) & (+0.064, +0.316, +0.947) & (+0.959, +0.247, −0.139) & (+0.401, +0.852, −0.335) \\
elbow (L+R) & (+0.879, +0.472, −0.072) & (+0.332, +0.925, −0.183) & (+0.270, −0.372, +0.888) \\
ankle (L+R) & (−0.279, +0.960, +0.007) & (+0.058, +0.296, +0.953) & (+0.780, −0.625, +0.033) \\
\end{longtable}

\begin{longtable}{@{}llrrrr@{}}
\caption{\textbf{Table A3.} Phase 3 projection weights and offsets per (joint, clinical joint angle). The three weights $(w_1, w_2, w_3)$ multiply the three swing-twist components and the offset $\delta$ is added; the per-side sign from Table A1 is multiplied into the input for bilateral joints. The full runtime formula is $\theta = w_1 c_0 + w_2 c_1 + w_3 c_2 + \delta$ (Section 2.2). Weights are fit with a ridge penalty (Section 2.3), which bounds their magnitude so extreme-range motion extrapolates smoothly; all weight vectors here have norm below 2.7. Bilateral joints share a single set of weights and offset.}
\label{tab:a3}\\
\toprule\noalign{}
Joint & Clinical angle & $w_1$ & $w_2$ & $w_3$ & $\delta$ (°) \\
\midrule\noalign{}
\endfirsthead
\toprule\noalign{}
Joint & Clinical angle & $w_1$ & $w_2$ & $w_3$ & $\delta$ (°) \\
\midrule\noalign{}
\endhead
\bottomrule\noalign{}
\endlastfoot
hip (L+R) & adduction & +0.596 & −0.024 & −0.062 & −1.397 \\
hip (L+R) & flexion & −0.075 & +1.060 & +0.137 & +11.752 \\
hip (L+R) & rotation & +0.137 & −0.002 & +0.596 & +1.503 \\
shoulder (L+R) & adduction & +0.512 & −0.003 & −0.052 & −50.288 \\
shoulder (L+R) & flexion & −0.028 & +1.163 & +0.034 & +34.220 \\
shoulder (L+R) & rotation & −0.175 & +0.029 & +0.816 & +21.581 \\
trunk & adduction & +0.623 & +0.016 & −0.343 & +1.449 \\
trunk & flexion & −0.082 & +0.844 & +0.364 & −6.645 \\
trunk & rotation & −0.029 & +0.042 & +0.470 & −1.575 \\
pelvis & flexion & +1.107 & −0.345 & +0.081 & −6.094 \\
pelvis & adduction & +0.007 & −0.122 & +0.224 & −0.883 \\
knee (L+R) & flexion & +0.753 & +0.139 & −0.991 & +2.902 \\
elbow (L+R) & flexion & −0.008 & +0.322 & +1.027 & +24.561 \\
ankle (L+R) & flexion & +0.049 & +0.863 & −0.529 & +6.698 \\
ankle (L+R) & adduction & +0.469 & −0.039 & +0.603 & −5.369 \\
\end{longtable}

Two structural observations are worth noting.

Sign corrections are not left-right symmetric for every channel: hip rotation is negated on one side but not the other, while the trunk and pelvis channels share a single sign. This reflects mirrored mediolateral conventions in the body model and is resolved by the calibration empirically rather than by anatomical assumption.

The shoulder offsets are large in magnitude (around −50° for adduction, +34° for flexion, and +22° for rotation) because the body model's rest pose is a T-pose with arms extended horizontally, while the clinical neutral position has the arms at the sides. These offsets reflect the geometric difference between the two convention's reference poses and not a deficiency of the extraction.

\section{Per-DOF results (GEM-X / SOMA)}

Full per-DOF results: held-out accuracy (Table B1), per-activity breakdown (Table B2), and per-activity cycle-aligned waveforms (Figures B1--B4). Per the validation protocol (Section 2.4); Table B1 is under the matched envelope used for the OpenCap Monocular comparison in Section 3.1.

\begin{longtable}{@{}
  >{\raggedright\arraybackslash}p{(\columnwidth - 6\tabcolsep) * \real{0.2800}}
  >{\raggedleft\arraybackslash}p{(\columnwidth - 6\tabcolsep) * \real{0.2400}}
  >{\raggedleft\arraybackslash}p{(\columnwidth - 6\tabcolsep) * \real{0.2400}}
  >{\raggedleft\arraybackslash}p{(\columnwidth - 6\tabcolsep) * \real{0.2400}}@{}}
\caption{\textbf{Table B1.} Per-DOF held-out accuracy: raw MAE, centred MAE (per-trial bias removed), Pearson r. n = 90 trials per channel under the matched envelope, aggregated across the nine LOSO folds (subjects 2, 3, 4, 5, 7, 8, 9, 10, 11; the complete OpenCap LabValidation cohort). Upper block: the fifteen DOFs directly comparable to OpenCap Monocular \cite{gilon2026opencapmono}; lower block: further channels we extract and report informally (shoulders, elbows, ankle inversion/eversion). A correlation in grey with a dagger ($^\dagger$) marks a cell whose ground-truth range of motion is below 15°: the per-trial correlation is statistically unreliable because the underlying signal is small, but raw and centred MAE are unaffected.}
\label{tab:b1}\\
\toprule\noalign{}
\begin{minipage}[b]{\linewidth}\raggedright
DOF
\end{minipage} & \begin{minipage}[b]{\linewidth}\raggedleft
MAE\_raw (°)
\end{minipage} & \begin{minipage}[b]{\linewidth}\raggedleft
MAE\_centred (°)
\end{minipage} & \begin{minipage}[b]{\linewidth}\raggedleft
r
\end{minipage} \\
\midrule\noalign{}
\endfirsthead
\toprule\noalign{}
\begin{minipage}[b]{\linewidth}\raggedright
DOF
\end{minipage} & \begin{minipage}[b]{\linewidth}\raggedleft
MAE\_raw (°)
\end{minipage} & \begin{minipage}[b]{\linewidth}\raggedleft
MAE\_centred (°)
\end{minipage} & \begin{minipage}[b]{\linewidth}\raggedleft
r
\end{minipage} \\
\midrule\noalign{}
\endhead
\bottomrule\noalign{}
\endlastfoot
Ankle flex/ext R & 4.84 & 3.18 & 0.807 \\
Ankle flex/ext L & 3.96 & 2.42 & 0.834 \\
Knee flex/ext R & 5.27 & 3.88 & 0.989 \\
Knee flex/ext L & 4.34 & 3.39 & 0.979 \\
Hip flex/ext R & 7.96 & 5.87 & 0.982 \\
Hip flex/ext L & 7.47 & 5.76 & 0.985 \\
Hip add/abd R & 4.00 & 2.56 & \textcolor{gray}{0.525}$^\dagger$ \\
Hip add/abd L & 3.93 & 1.94 & \textcolor{gray}{0.770}$^\dagger$ \\
Hip rotation R & 3.80 & 2.41 & \textcolor{gray}{0.335}$^\dagger$ \\
Hip rotation L & 4.27 & 3.14 & \textcolor{gray}{0.494}$^\dagger$ \\
Pelvis tilt & 4.79 & 3.38 & 0.491 \\
Pelvis list & 1.59 & 1.14 & \textcolor{gray}{0.485}$^\dagger$ \\
Trunk flex/ext (lumbar proxy) & 5.75 & 4.14 & 0.555 \\
Trunk lateral bend (lumbar proxy) & 2.56 & 1.57 & 0.600 \\
Trunk rotation (lumbar proxy) & 2.94 & 1.88 & \textcolor{gray}{0.598}$^\dagger$ \\
\textbf{15-DOF mean (OpenCap-matched)} & \textbf{4.50} & \textbf{3.11} & \textbf{0.695} \\
Shoulder flex/ext R & 6.19 & 4.12 & 0.931 \\
Shoulder flex/ext L & 5.72 & 3.68 & 0.895 \\
Shoulder abd/add R & 4.56 & 2.73 & \textcolor{gray}{0.313}$^\dagger$ \\
Shoulder abd/add L & 4.99 & 3.37 & 0.588 \\
Shoulder rotation R & 10.56 & 4.37 & 0.417 \\
Shoulder rotation L & 12.86 & 4.47 & 0.249 \\
Elbow flex/ext R & 5.89 & 3.17 & 0.816 \\
Elbow flex/ext L & 6.55 & 4.25 & 0.788 \\
Ankle inv/ev R & 5.20 & 3.57 & 0.316 \\
Ankle inv/ev L & 6.85 & 3.33 & 0.077 \\
\textbf{25-DOF mean (all extracted channels)} & \textbf{5.47} & \textbf{3.35} & \textbf{0.63} \\
\end{longtable}

{\footnotesize
\begin{longtable}{@{}
  >{\raggedright\arraybackslash}p{(\columnwidth - 8\tabcolsep) * \real{0.2800}}
  >{\raggedright\arraybackslash}p{(\columnwidth - 8\tabcolsep) * \real{0.1800}}
  >{\raggedright\arraybackslash}p{(\columnwidth - 8\tabcolsep) * \real{0.1800}}
  >{\raggedright\arraybackslash}p{(\columnwidth - 8\tabcolsep) * \real{0.1800}}
  >{\raggedright\arraybackslash}p{(\columnwidth - 8\tabcolsep) * \real{0.1800}}@{}}
\caption{\textbf{Table B2.} Per DOF × activity accuracy, held-out LOSO, pooled across all 9 subjects and all 5 camera views (n = 270 trials per activity for walking and drop\_jump, 90 each for squats and sit-to-stand). Each cell is \textbf{r / RMSE° / MAE°}, same per-trial alignment as Table B1. Low-signal $^\dagger$ convention as in Table B1.}
\label{tab:b2}\\
\toprule\noalign{}
\begin{minipage}[b]{\linewidth}\raggedright
DOF
\end{minipage} & \begin{minipage}[b]{\linewidth}\raggedright
walking
\end{minipage} & \begin{minipage}[b]{\linewidth}\raggedright
drop\_jump
\end{minipage} & \begin{minipage}[b]{\linewidth}\raggedright
squats
\end{minipage} & \begin{minipage}[b]{\linewidth}\raggedright
sit-to-stand
\end{minipage} \\
\midrule\noalign{}
\endfirsthead
\toprule\noalign{}
\begin{minipage}[b]{\linewidth}\raggedright
DOF
\end{minipage} & \begin{minipage}[b]{\linewidth}\raggedright
walking
\end{minipage} & \begin{minipage}[b]{\linewidth}\raggedright
drop\_jump
\end{minipage} & \begin{minipage}[b]{\linewidth}\raggedright
squats
\end{minipage} & \begin{minipage}[b]{\linewidth}\raggedright
sit-to-stand
\end{minipage} \\
\midrule\noalign{}
\endhead
\bottomrule\noalign{}
\endlastfoot
& {\scriptsize\itshape r / RMSE° / MAE°} & {\scriptsize\itshape r / RMSE° / MAE°} & {\scriptsize\itshape r / RMSE° / MAE°} & {\scriptsize\itshape r / RMSE° / MAE°} \\
Ankle flex/ext R & 0.77 / 6.9 / 5.3 & 0.96 / 11.2 / 8.3 & 0.96 / 5.7 / 4.2 & 0.79 / 7.2 / 5.3 \\
Ankle flex/ext L & 0.76 / 7.3 / 5.5 & 0.96 / 11.5 / 8.5 & 0.94 / 5.9 / 4.4 & 0.79 / 6.6 / 4.8 \\
Knee flex/ext R & 0.98 / 6.2 / 4.9 & 0.97 / 11.3 / 8.0 & 0.98 / 10.7 / 6.7 & 0.93 / 15.0 / 8.4 \\
Knee flex/ext L & 0.97 / 7.1 / 5.0 & 0.97 / 10.8 / 7.5 & 0.98 / 11.5 / 6.3 & 0.93 / 15.2 / 8.2 \\
Hip flex/ext R & 0.98 / 8.5 / 7.2 & 0.96 / 9.0 / 6.5 & 0.99 / 13.0 / 9.4 & 0.93 / 14.1 / 10.0 \\
Hip flex/ext L & 0.97 / 6.6 / 4.8 & 0.96 / 9.3 / 6.5 & 0.98 / 11.0 / 8.5 & 0.94 / 14.1 / 9.7 \\
Hip add/abd R & 0.79 / 5.8 / 4.9 & \textcolor{gray}{0.13}$^\dagger$ / 5.1 / 3.9 & \textcolor{gray}{0.48}$^\dagger$ / 6.2 / 4.7 & \textcolor{gray}{-0.20}$^\dagger$ / 4.7 / 3.7 \\
Hip add/abd L & 0.85 / 4.2 / 3.5 & \textcolor{gray}{0.41}$^\dagger$ / 4.8 / 3.7 & 0.80 / 7.1 / 5.3 & \textcolor{gray}{0.44}$^\dagger$ / 4.3 / 3.2 \\
Hip rotation R & \textcolor{gray}{0.26}$^\dagger$ / 4.9 / 3.9 & 0.44 / 6.2 / 4.7 & \textcolor{gray}{0.30}$^\dagger$ / 5.7 / 4.6 & \textcolor{gray}{0.52}$^\dagger$ / 5.1 / 3.8 \\
Hip rotation L & \textcolor{gray}{0.22}$^\dagger$ / 5.6 / 4.4 & 0.73 / 6.2 / 4.6 & 0.88 / 7.3 / 5.7 & 0.59 / 4.8 / 3.6 \\
Pelvis tilt & \textcolor{gray}{0.22}$^\dagger$ / 4.5 / 3.7 & 0.78 / 7.0 / 5.0 & 0.94 / 6.7 / 4.6 & 0.91 / 8.8 / 6.1 \\
Pelvis list & \textcolor{gray}{0.78}$^\dagger$ / 3.2 / 2.5 & \textcolor{gray}{0.24}$^\dagger$ / 3.3 / 2.5 & \textcolor{gray}{0.05}$^\dagger$ / 2.4 / 1.8 & \textcolor{gray}{-0.01}$^\dagger$ / 1.6 / 1.3 \\
Trunk flex/ext (lumbar proxy) & \textcolor{gray}{0.26}$^\dagger$ / 4.1 / 3.3 & 0.90 / 6.2 / 4.7 & 0.95 / 7.3 / 5.6 & 0.81 / 10.4 / 7.7 \\
Trunk lateral bend (lumbar proxy) & 0.79 / 7.7 / 6.0 & \textcolor{gray}{0.23}$^\dagger$ / 3.7 / 2.9 & \textcolor{gray}{0.20}$^\dagger$ / 4.0 / 2.9 & \textcolor{gray}{0.31}$^\dagger$ / 2.8 / 1.8 \\
Trunk rotation (lumbar proxy) & 0.84 / 5.5 / 4.3 & 0.21 / 6.7 / 4.8 & \textcolor{gray}{0.28}$^\dagger$ / 4.1 / 2.8 & \textcolor{gray}{0.09}$^\dagger$ / 3.8 / 2.8 \\
Shoulder flex/ext R & 0.83 / 6.7 / 4.9 & 0.75 / 41.6 / 27.0 & 0.98 / 9.2 / 6.8 & 0.88 / 13.2 / 8.4 \\
Shoulder flex/ext L & 0.86 / 7.3 / 6.0 & 0.73 / 43.7 / 27.3 & 0.98 / 9.2 / 6.4 & 0.88 / 12.8 / 8.1 \\
Shoulder abd/add R & \textcolor{gray}{0.50}$^\dagger$ / 5.7 / 4.5 & 0.55 / 22.7 / 13.3 & 0.45 / 6.3 / 4.7 & 0.40 / 6.8 / 5.2 \\
Shoulder abd/add L & \textcolor{gray}{0.56}$^\dagger$ / 5.2 / 4.2 & 0.50 / 27.7 / 14.9 & 0.34 / 6.7 / 5.5 & 0.36 / 6.5 / 5.1 \\
Shoulder rotation R & 0.27 / 16.3 / 14.1 & 0.71 / 35.0 / 21.5 & 0.64 / 13.0 / 10.7 & 0.60 / 11.7 / 9.2 \\
Shoulder rotation L & 0.31 / 18.1 / 15.3 & 0.61 / 40.4 / 24.4 & 0.44 / 16.0 / 13.5 & 0.65 / 14.3 / 11.7 \\
Elbow flex/ext R & 0.75 / 10.1 / 6.9 & 0.76 / 13.5 / 10.3 & 0.79 / 9.8 / 7.4 & 0.68 / 10.3 / 6.6 \\
Elbow flex/ext L & 0.82 / 8.7 / 7.0 & 0.72 / 13.7 / 10.5 & 0.78 / 9.7 / 7.8 & 0.68 / 10.8 / 7.6 \\
Ankle inv/ev R & 0.17 / 9.7 / 7.9 & 0.51 / 10.2 / 8.4 & 0.37 / 7.9 / 6.2 & 0.23 / 6.0 / 4.8 \\
Ankle inv/ev L & 0.02 / 10.2 / 8.3 & 0.37 / 8.8 / 7.1 & 0.05 / 9.4 / 7.6 & 0.30 / 7.9 / 5.7 \\
\textbf{25-DOF mean} & 0.62 / 7.4 / 5.9 & 0.64 / 14.8 / 9.9 & 0.66 / 8.2 / 6.2 & 0.58 / 8.7 / 6.1 \\
\end{longtable}
}

\subsection*{Per-DOF waveforms across activities}

Figures B1--B4 show per-DOF angle waveforms, one figure per activity. Each subplot overlays predicted (orange) on reference (black); solid line = mean across cycles, shaded band = ±1 SD. Cam1 throughout, matching Section 3.1.

For walking, squats, and sit-to-stand, repeating cycles are detected from peaks in the reference knee-flexion signal and time-normalised to 0--100\%. Trials short enough to contain just one cycle (most walking trials) are treated as one cycle. Drop-jump trials are time-normalised to 0--100\% of trial duration.

Subplot titles list r, MAE, and cMAE pooled across the included cycles (n = cycle count). These metrics use the production calibration applied to the cohort and are in-sample with respect to it; the Appendix B tables are LOSO held-out and so slightly more conservative.

\begin{figure}[H]
  \centering
  \includegraphics[width=\textwidth]{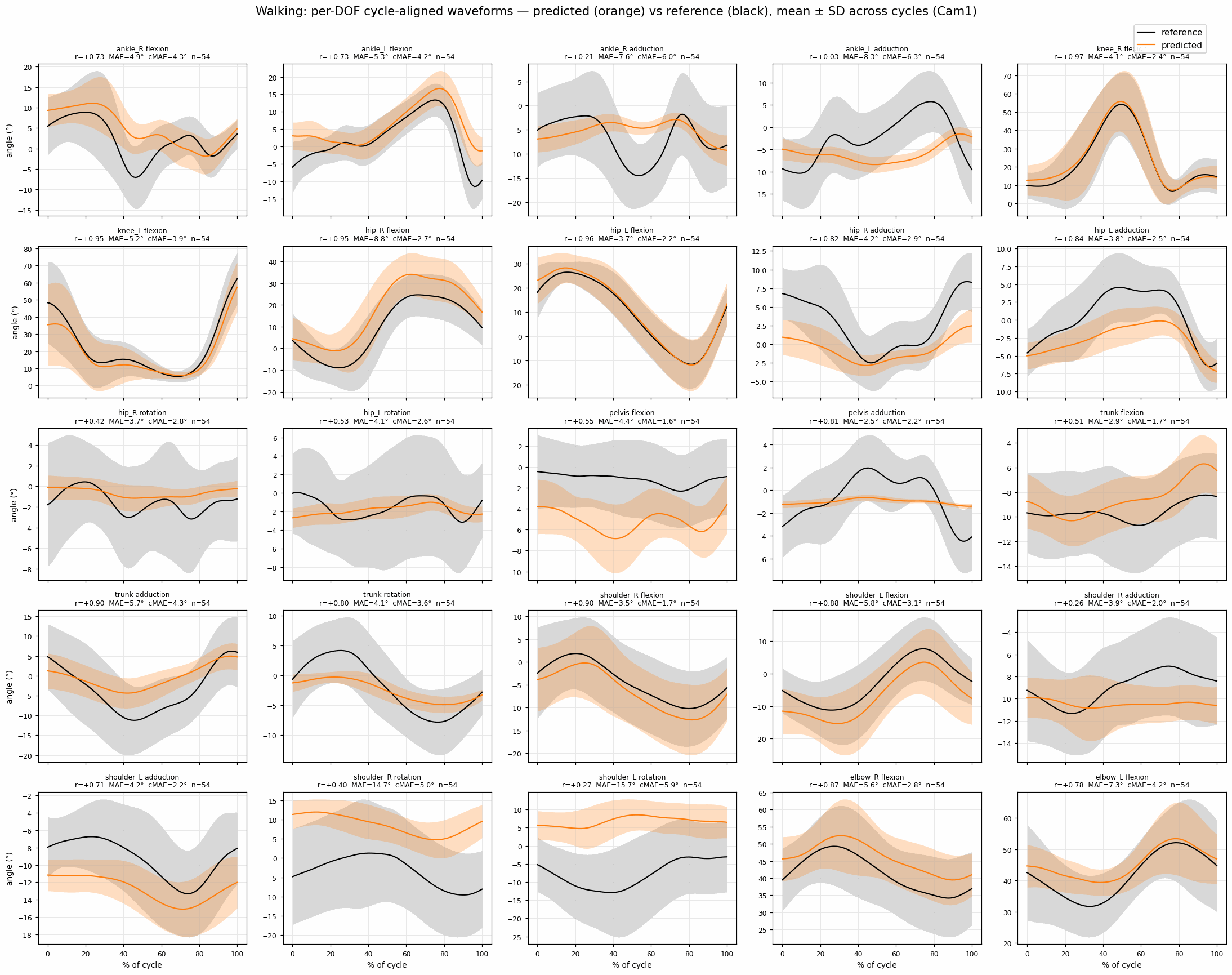}
  \caption{Walking: per-DOF joint-angle waveforms across the gait cycle (angle in degrees, x-axis 0--100\% of cycle), predicted (orange) over the OpenSim reference (black), mean ±1 SD across cycles. Close overlap means accurate waveform tracking.}
  \label{fig:waveforms_walking}
\end{figure}

\begin{figure}[H]
  \centering
  \includegraphics[width=\textwidth]{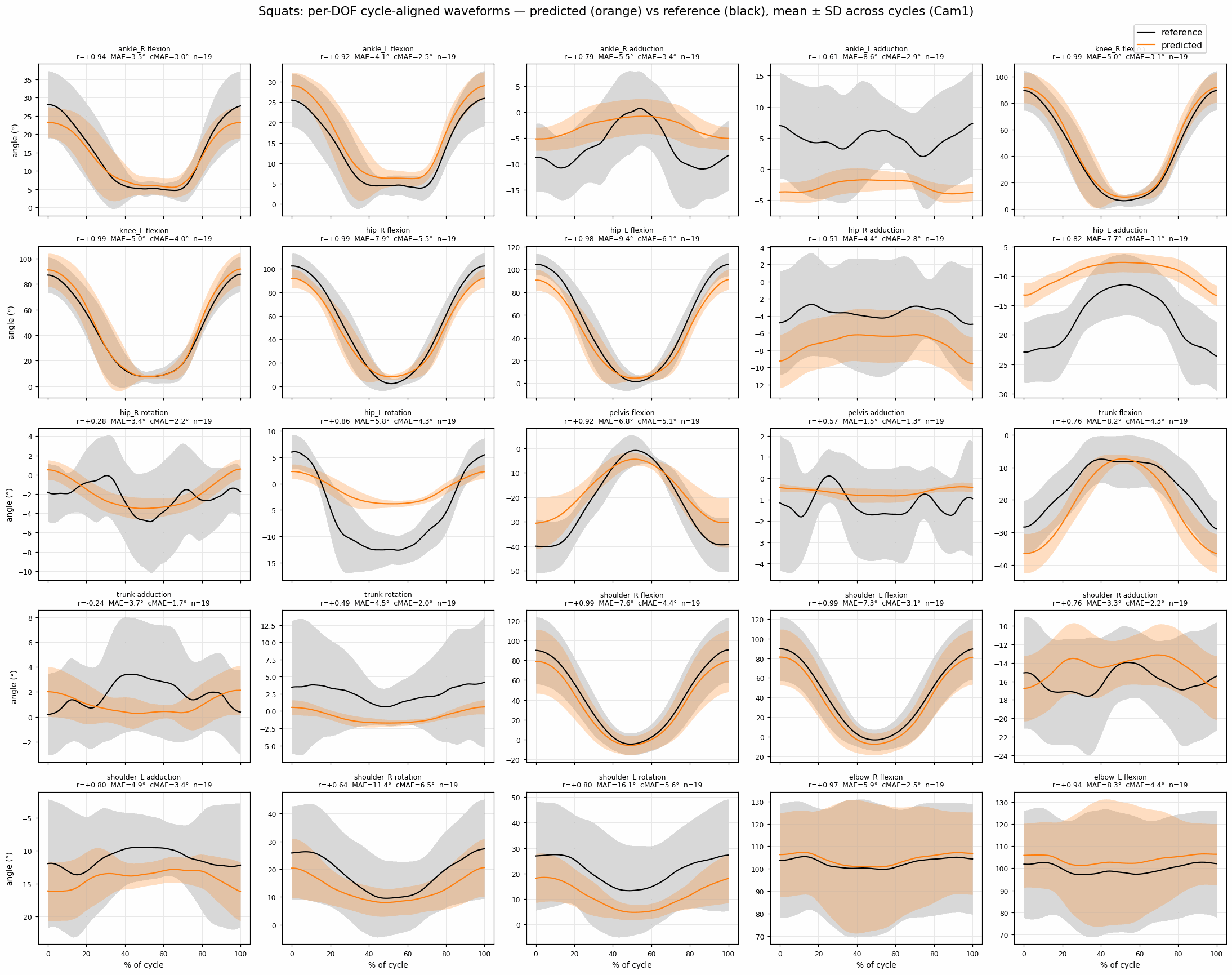}
  \caption{Squats: per-DOF joint-angle waveforms across the movement cycle (degrees vs 0--100\% of cycle), predicted (orange) over reference (black), mean ±1 SD across cycles.}
  \label{fig:waveforms_squats}
\end{figure}

\begin{figure}[H]
  \centering
  \includegraphics[width=\textwidth]{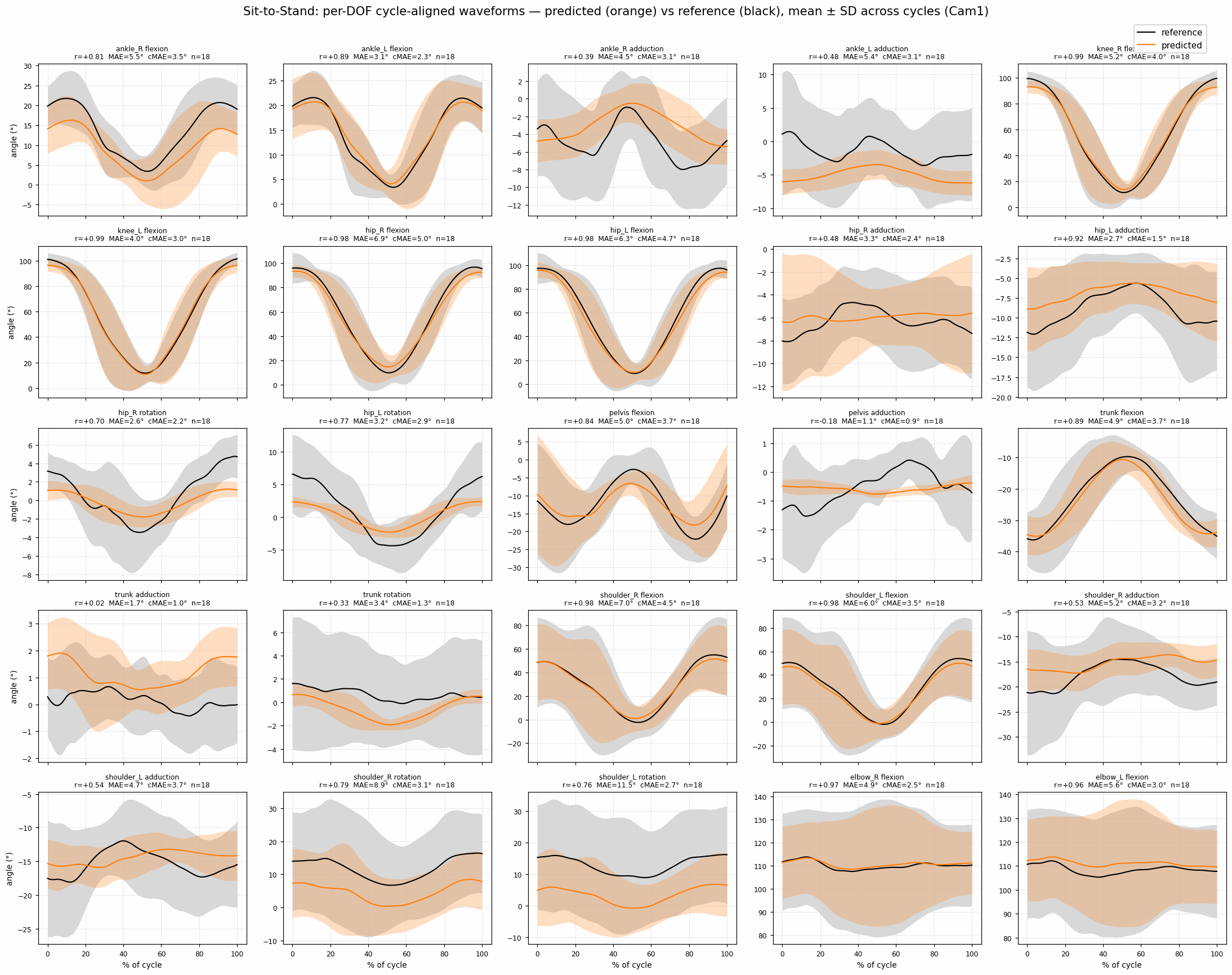}
  \caption{Sit-to-stand: per-DOF joint-angle waveforms across the movement cycle (degrees vs 0--100\% of cycle), predicted (orange) over reference (black), mean ±1 SD across cycles.}
  \label{fig:waveforms_sit_to_stand}
\end{figure}

\begin{figure}[H]
  \centering
  \includegraphics[width=\textwidth]{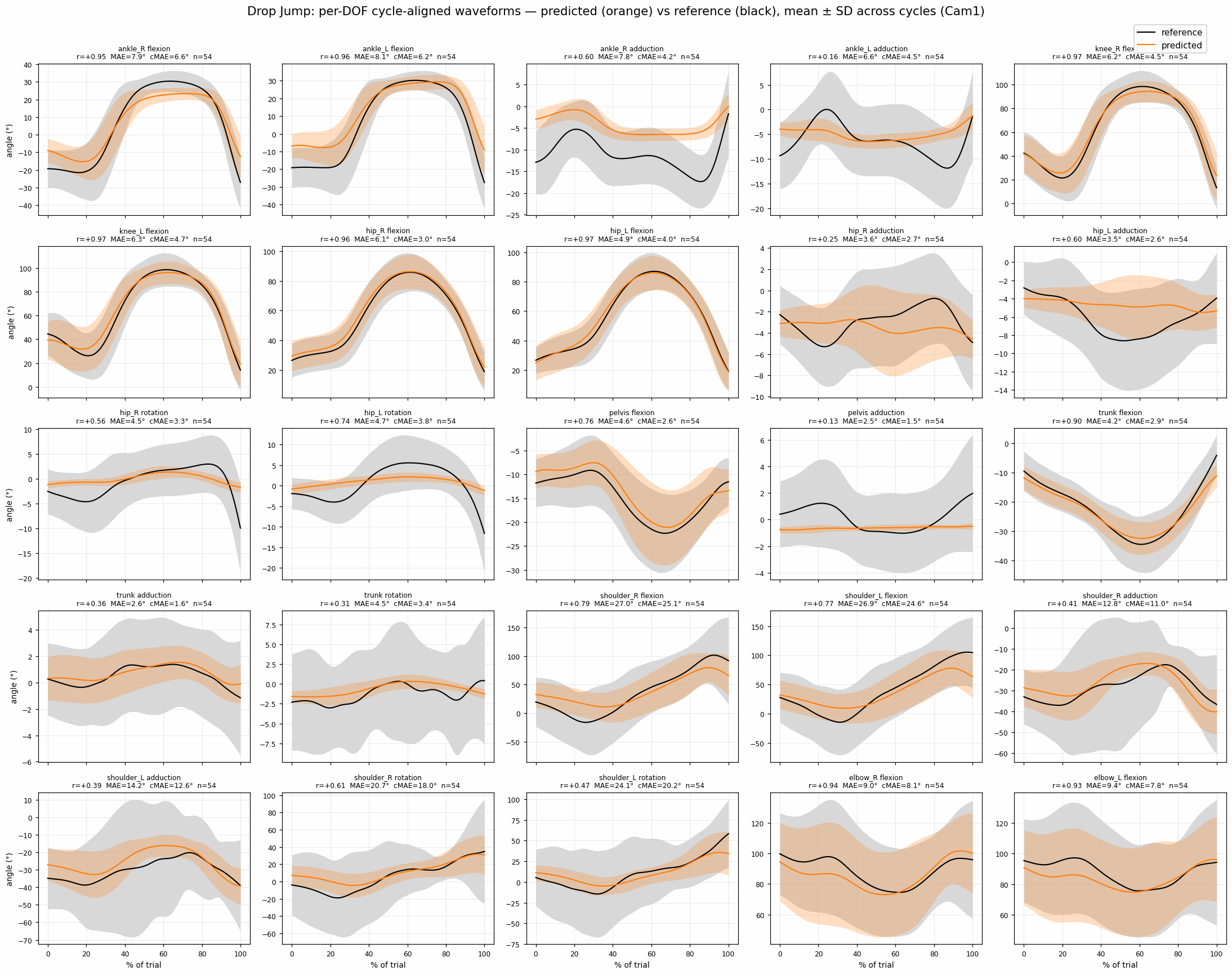}
  \caption{Drop jumps: per-DOF joint-angle waveforms (degrees vs 0--100\% of trial duration), predicted (orange) over reference (black), mean ±1 SD. These trials sit outside the validated envelope of Section 3.1 and are shown for completeness, so the envelope-exclusion argument can be inspected visually rather than only in aggregate.}
  \label{fig:waveforms_drop_jump}
\end{figure}

\section{Cross-body-model benchmark --- SAM 3D Body (MHR) vs GEM-X (SOMA)}
\subsection{Setup}

\begin{itemize}
\item
  \textbf{Estimators.} SAM 3D Body \cite{yang2026sam3d} on the MHR skeleton (127 joints) \cite{ferguson2025mhr} vs the paper's GEM-X on SOMA (77 joints), both from monocular video.
\item
  \textbf{Only the body model changes.} The Section 2.2 extraction and Section 2.3 calibration are identical; the sole change for SAM 3D Body is a regenerated calibration table.
\item
  \textbf{Protocol.} Per Section 2.4: OpenCap LabValidation, OpenSim IK reference (LaiUhlrich2022, 100 Hz), the same metrics (raw MAE, centred MAE, Pearson r), leave-one-subject-out.
\item
  \textbf{Matched envelope.} Both columns: single oblique camera (Cam1), non-jumping activities, 15 OpenCap-comparable DOFs, n = 90 held-out trials per channel.
\end{itemize}

\subsection{Runtime}

The extraction adds about a millisecond per frame regardless of estimator, so the body model sets the frame rate, not the method. The two estimators sit at opposite ends: GEM-X is an offline, accuracy-first batch model, whereas SAM 3D Body is built for \textbf{real-time use} --- its native engine targets live single-camera capture (interactive demos, webcam-driven feedback, kiosk gait screening). Achieved throughput depends heavily on the GPU and the inference backend, and reaching the highest rates is not turnkey: it needs a tuned engine build rather than the stock runtime. We therefore describe SAM 3D Body as real-time \emph{capable} on commodity hardware rather than quoting a single number. The relevant conclusion for this paper is that the extraction imposes no real-time cost; whether a deployment runs offline or live is purely an estimator choice.

\subsection{Per-DOF held-out accuracy}

Per the validation protocol (Section 2.4). The GEM-X column reproduces Table B1 exactly. SAM 3D Body parameters come from the \textbf{same raw TensorRT-FP16 inference path used in real-time deployment} (Section C.2), so these are the deployed numbers, not a research-only configuration.

\begin{longtable}{@{}
  >{\raggedright\arraybackslash}p{(\columnwidth - 12\tabcolsep) * \real{0.2200}}
  >{\raggedleft\arraybackslash}p{(\columnwidth - 12\tabcolsep) * \real{0.1300}}
  >{\raggedleft\arraybackslash}p{(\columnwidth - 12\tabcolsep) * \real{0.1300}}
  >{\raggedleft\arraybackslash}p{(\columnwidth - 12\tabcolsep) * \real{0.1300}}
  >{\raggedleft\arraybackslash}p{(\columnwidth - 12\tabcolsep) * \real{0.1300}}
  >{\raggedleft\arraybackslash}p{(\columnwidth - 12\tabcolsep) * \real{0.1300}}
  >{\raggedleft\arraybackslash}p{(\columnwidth - 12\tabcolsep) * \real{0.1300}}@{}}
\caption{\textbf{Table C1.} Same column layout as Table B1 (raw MAE, centred MAE, Pearson r), one estimator block beside the other. The GEM-X columns are byte-identical to Table B1. Low-signal $^\dagger$ convention as in Table B1; the same channels are flagged in both columns since that property comes from the reference signal.}
\label{tab:c1}\\
\toprule\noalign{}
\begin{minipage}[b]{\linewidth}\raggedright
DOF
\end{minipage} & \begin{minipage}[b]{\linewidth}\raggedleft
GEM-X MAE\_raw (°)
\end{minipage} & \begin{minipage}[b]{\linewidth}\raggedleft
GEM-X MAE\_cen (°)
\end{minipage} & \begin{minipage}[b]{\linewidth}\raggedleft
GEM-X r
\end{minipage} & \begin{minipage}[b]{\linewidth}\raggedleft
SAM3D MAE\_raw (°)
\end{minipage} & \begin{minipage}[b]{\linewidth}\raggedleft
SAM3D MAE\_cen (°)
\end{minipage} & \begin{minipage}[b]{\linewidth}\raggedleft
SAM3D r
\end{minipage} \\
\midrule\noalign{}
\endfirsthead
\toprule\noalign{}
\begin{minipage}[b]{\linewidth}\raggedright
DOF
\end{minipage} & \begin{minipage}[b]{\linewidth}\raggedleft
GEM-X MAE\_raw (°)
\end{minipage} & \begin{minipage}[b]{\linewidth}\raggedleft
GEM-X MAE\_cen (°)
\end{minipage} & \begin{minipage}[b]{\linewidth}\raggedleft
GEM-X r
\end{minipage} & \begin{minipage}[b]{\linewidth}\raggedleft
SAM3D MAE\_raw (°)
\end{minipage} & \begin{minipage}[b]{\linewidth}\raggedleft
SAM3D MAE\_cen (°)
\end{minipage} & \begin{minipage}[b]{\linewidth}\raggedleft
SAM3D r
\end{minipage} \\
\midrule\noalign{}
\endhead
\bottomrule\noalign{}
\endlastfoot
Ankle flex/ext R & 4.84 & 3.18 & 0.807 & 5.67 & 4.56 & \textcolor{gray}{0.133}$^\dagger$ \\
Ankle flex/ext L & 3.96 & 2.42 & 0.834 & 6.79 & 5.92 & 0.467 \\
Knee flex/ext R & 5.27 & 3.88 & 0.989 & 4.27 & 3.10 & 0.991 \\
Knee flex/ext L & 4.34 & 3.39 & 0.979 & 4.86 & 3.26 & 0.979 \\
Hip flex/ext R & 7.96 & 5.87 & 0.982 & 7.39 & 5.91 & 0.986 \\
Hip flex/ext L & 7.47 & 5.76 & 0.985 & 6.10 & 5.38 & 0.985 \\
Hip add/abd R & 4.00 & 2.56 & \textcolor{gray}{0.525}$^\dagger$ & 3.41 & 2.27 & \textcolor{gray}{0.528}$^\dagger$ \\
Hip add/abd L & 3.93 & 1.94 & \textcolor{gray}{0.770}$^\dagger$ & 4.38 & 2.63 & \textcolor{gray}{0.654}$^\dagger$ \\
Hip rotation R & 3.80 & 2.41 & \textcolor{gray}{0.335}$^\dagger$ & 3.77 & 2.50 & \textcolor{gray}{0.296}$^\dagger$ \\
Hip rotation L & 4.27 & 3.14 & \textcolor{gray}{0.494}$^\dagger$ & 3.96 & 2.77 & \textcolor{gray}{0.411}$^\dagger$ \\
Pelvis tilt & 4.79 & 3.38 & 0.491 & 7.29 & 5.33 & 0.564 \\
Pelvis list & 1.59 & 1.14 & \textcolor{gray}{0.485}$^\dagger$ & 2.45 & 1.30 & \textcolor{gray}{0.229}$^\dagger$ \\
Trunk flex/ext (lumbar proxy) & 5.75 & 4.14 & 0.555 & 4.69 & 3.65 & 0.714 \\
Trunk lateral bend (lumbar proxy) & 2.56 & 1.57 & 0.600 & 2.12 & 1.40 & 0.684 \\
Trunk rotation (lumbar proxy) & 2.94 & 1.88 & \textcolor{gray}{0.598}$^\dagger$ & 2.82 & 1.90 & \textcolor{gray}{0.604}$^\dagger$ \\
\textbf{15-DOF mean (OpenCap-matched)} & \textbf{4.50} & \textbf{3.11} & \textbf{0.695} & \textbf{4.66} & \textbf{3.46} & \textbf{0.615} \\
\end{longtable}

\subsection{Reading}

Both body models land in the same accuracy range on raw MAE: SAM 3D Body \textbf{4.66°} against GEM-X \textbf{4.50°}. Resampling the nine subjects (bootstrap, 20,000 draws, the §3.1 method) gives SAM 3D Body a 95\% CI of \textbf{4.37°--5.19°} and GEM-X \textbf{4.28°--4.78°}; the \emph{paired} per-subject difference is \textbf{+0.20° (95\% CI −0.05° to +0.48°)}, whose interval spans zero, so the two skeletons are statistically indistinguishable on this cohort. The same closeness holds on centred MAE: \textbf{3.46° vs 3.11°}, about 0.35° in GEM-X's favour --- a small residual consistent with the architectures, since GEM-X (GENMO) is a diffusion model with a temporal prior whereas SAM 3D Body is a single-pass feed-forward estimator with no temporal model and so carries slightly more per-frame jitter.

Channel by channel the two skeletons trade places within a narrow band: SAM 3D Body tracks the knee marginally tighter (4.3--4.9° vs 4.3--5.3°) and trunk flexion better (4.7° vs 5.8°), while GEM-X is better on pelvis tilt (4.8° vs 7.3°). The strong/weak split is shared: high-signal flexion (knee r 0.98--0.99, hip r 0.98--0.99) tracks tightly, the out-of-plane channels (ankle, hip rotation, pelvis list) are the noisy ones --- consistent with a property of monocular sensing rather than of any particular body model.

\subsection{Per-activity breakdown}

Broken out per activity across all five cameras (SAM 3D Body / MHR). The flexion channels stay strong; drop jumps are the hardest case (fast landings, motion blur, all-camera pooling), and the all-camera ankle estimate is the weak channel.

{\footnotesize
\begin{longtable}{@{}
  >{\raggedright\arraybackslash}p{(\columnwidth - 8\tabcolsep) * \real{0.2800}}
  >{\raggedright\arraybackslash}p{(\columnwidth - 8\tabcolsep) * \real{0.1800}}
  >{\raggedright\arraybackslash}p{(\columnwidth - 8\tabcolsep) * \real{0.1800}}
  >{\raggedright\arraybackslash}p{(\columnwidth - 8\tabcolsep) * \real{0.1800}}
  >{\raggedright\arraybackslash}p{(\columnwidth - 8\tabcolsep) * \real{0.1800}}@{}}
\caption{\textbf{Table C2.} Same layout as Table B2: one row per DOF, activities as columns, each cell \textbf{r / RMSE° / MAE°}, pooled across all five camera views. Low-signal $^\dagger$ convention as in Table B1.}
\label{tab:c2}\\
\toprule\noalign{}
\begin{minipage}[b]{\linewidth}\raggedright
DOF
\end{minipage} & \begin{minipage}[b]{\linewidth}\raggedright
walking
\end{minipage} & \begin{minipage}[b]{\linewidth}\raggedright
drop\_jump
\end{minipage} & \begin{minipage}[b]{\linewidth}\raggedright
squats
\end{minipage} & \begin{minipage}[b]{\linewidth}\raggedright
sit-to-stand
\end{minipage} \\
\midrule\noalign{}
\endfirsthead
\toprule\noalign{}
\begin{minipage}[b]{\linewidth}\raggedright
DOF
\end{minipage} & \begin{minipage}[b]{\linewidth}\raggedright
walking
\end{minipage} & \begin{minipage}[b]{\linewidth}\raggedright
drop\_jump
\end{minipage} & \begin{minipage}[b]{\linewidth}\raggedright
squats
\end{minipage} & \begin{minipage}[b]{\linewidth}\raggedright
sit-to-stand
\end{minipage} \\
\midrule\noalign{}
\endhead
\bottomrule\noalign{}
\endlastfoot
& {\scriptsize\itshape r / RMSE° / MAE°} & {\scriptsize\itshape r / RMSE° / MAE°} & {\scriptsize\itshape r / RMSE° / MAE°} & {\scriptsize\itshape r / RMSE° / MAE°} \\
Ankle flex/ext R & -0.21 / 11.9 / 9.5 & 0.78 / 24.6 / 18.8 & 0.94 / 7.3 / 5.8 & 0.76 / 7.0 / 5.4 \\
Ankle flex/ext L & -0.20 / 10.5 / 8.5 & 0.67 / 22.8 / 19.8 & 0.88 / 11.9 / 9.0 & 0.44 / 7.3 / 5.8 \\
Knee flex/ext R & 0.98 / 6.6 / 5.0 & 0.99 / 8.7 / 6.6 & 1.00 / 7.3 / 5.9 & 1.00 / 5.8 / 4.6 \\
Knee flex/ext L & 0.96 / 7.5 / 5.2 & 0.99 / 7.4 / 5.7 & 1.00 / 6.5 / 5.2 & 1.00 / 6.1 / 5.0 \\
Hip flex/ext R & 0.98 / 10.2 / 8.4 & 0.98 / 8.8 / 6.9 & 0.99 / 7.7 / 6.2 & 0.98 / 10.0 / 8.0 \\
Hip flex/ext L & 0.98 / 10.2 / 8.9 & 0.98 / 8.0 / 6.4 & 1.00 / 8.1 / 6.4 & 0.98 / 8.9 / 7.1 \\
Hip add/abd R & 0.78 / 3.7 / 3.0 & \textcolor{gray}{0.28}$^\dagger$ / 5.0 / 3.9 & \textcolor{gray}{0.47}$^\dagger$ / 4.9 / 3.7 & \textcolor{gray}{0.12}$^\dagger$ / 4.2 / 3.3 \\
Hip add/abd L & \textcolor{gray}{0.65}$^\dagger$ / 4.7 / 3.7 & \textcolor{gray}{0.52}$^\dagger$ / 4.4 / 3.5 & 0.70 / 5.6 / 4.1 & \textcolor{gray}{0.39}$^\dagger$ / 4.1 / 3.1 \\
Hip rotation R & \textcolor{gray}{0.28}$^\dagger$ / 5.1 / 4.1 & 0.45 / 5.7 / 4.4 & \textcolor{gray}{0.40}$^\dagger$ / 4.9 / 3.9 & \textcolor{gray}{0.68}$^\dagger$ / 3.7 / 3.1 \\
Hip rotation L & \textcolor{gray}{0.22}$^\dagger$ / 5.3 / 4.2 & 0.73 / 5.1 / 3.9 & 0.90 / 5.4 / 4.3 & 0.74 / 3.9 / 3.1 \\
Pelvis tilt & \textcolor{gray}{0.33}$^\dagger$ / 9.7 / 8.5 & 0.81 / 4.8 / 3.8 & 0.87 / 7.8 / 6.3 & 0.88 / 8.2 / 6.6 \\
Pelvis list & \textcolor{gray}{0.35}$^\dagger$ / 4.3 / 3.4 & \textcolor{gray}{0.34}$^\dagger$ / 4.1 / 3.3 & \textcolor{gray}{0.20}$^\dagger$ / 3.8 / 3.1 & \textcolor{gray}{0.14}$^\dagger$ / 4.2 / 3.4 \\
Trunk flex/ext (lumbar proxy) & \textcolor{gray}{0.54}$^\dagger$ / 3.1 / 2.6 & 0.95 / 4.6 / 3.4 & 0.96 / 5.2 / 4.0 & 0.93 / 7.1 / 5.5 \\
Trunk lateral bend (lumbar proxy) & 0.86 / 5.0 / 3.9 & \textcolor{gray}{0.45}$^\dagger$ / 3.9 / 3.1 & \textcolor{gray}{0.45}$^\dagger$ / 3.0 / 2.3 & \textcolor{gray}{0.37}$^\dagger$ / 2.3 / 1.8 \\
Trunk rotation (lumbar proxy) & 0.85 / 4.8 / 3.8 & 0.32 / 6.4 / 4.5 & \textcolor{gray}{0.21}$^\dagger$ / 4.0 / 2.8 & \textcolor{gray}{0.24}$^\dagger$ / 3.5 / 2.7 \\
\end{longtable}
}

\subsection{Cycle-aligned waveforms}

Per-activity cycle-aligned predicted-vs-reference waveforms across the full per-DOF grid, laid out 1-to-1 with Appendix B; predicted in orange, reference in black, mean ± SD across cycles on Cam1.

\begin{figure}[H]
  \centering
  \includegraphics[width=\textwidth]{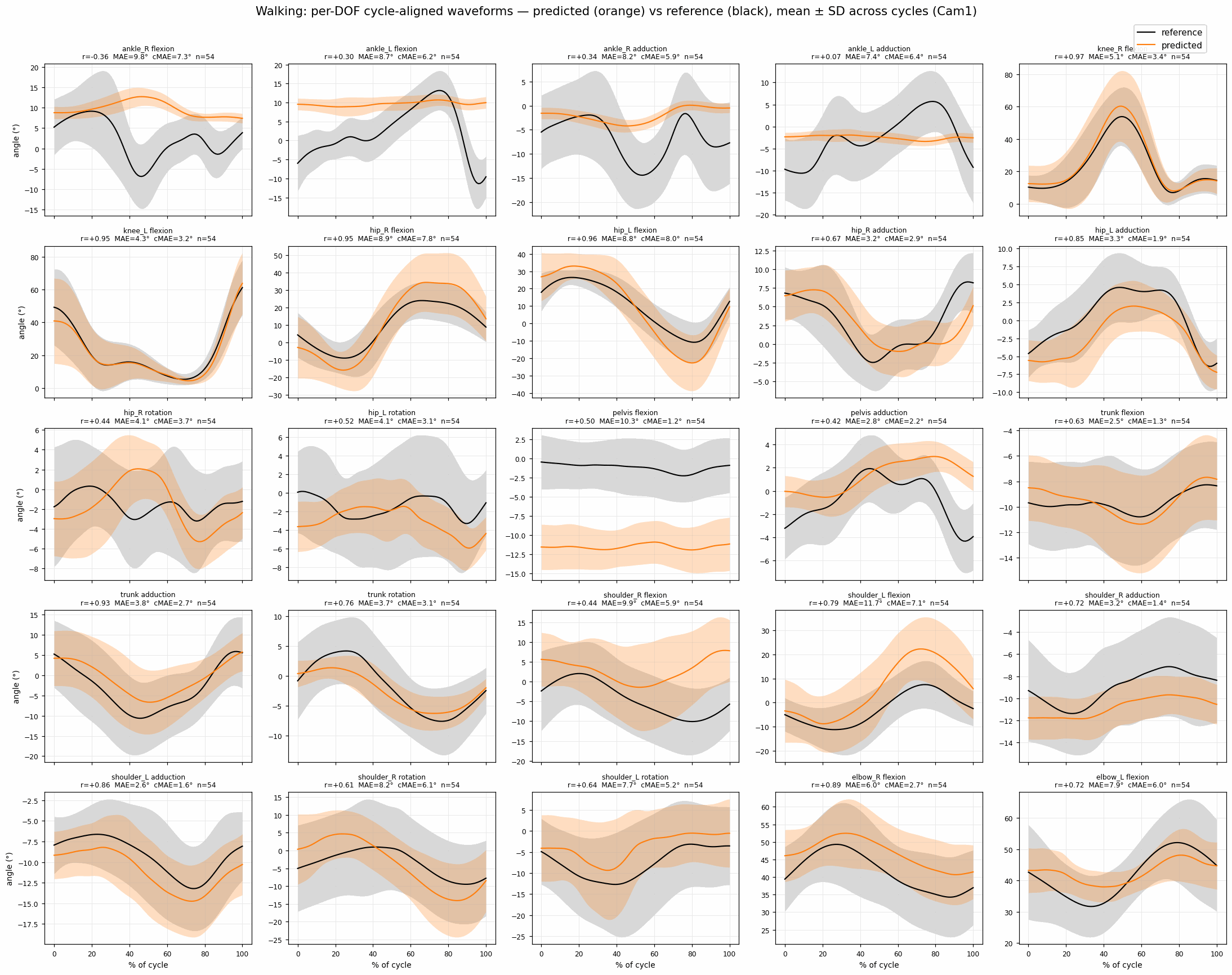}
  \caption{SAM 3D Body, walking: per-DOF joint-angle waveforms (degrees vs 0--100\% of gait cycle), predicted (orange) over reference (black), mean ±1 SD across cycles. Same layout as Figure B1.}
  \label{fig:sam3dbody_waveforms_walking}
\end{figure}

\begin{figure}[H]
  \centering
  \includegraphics[width=\textwidth]{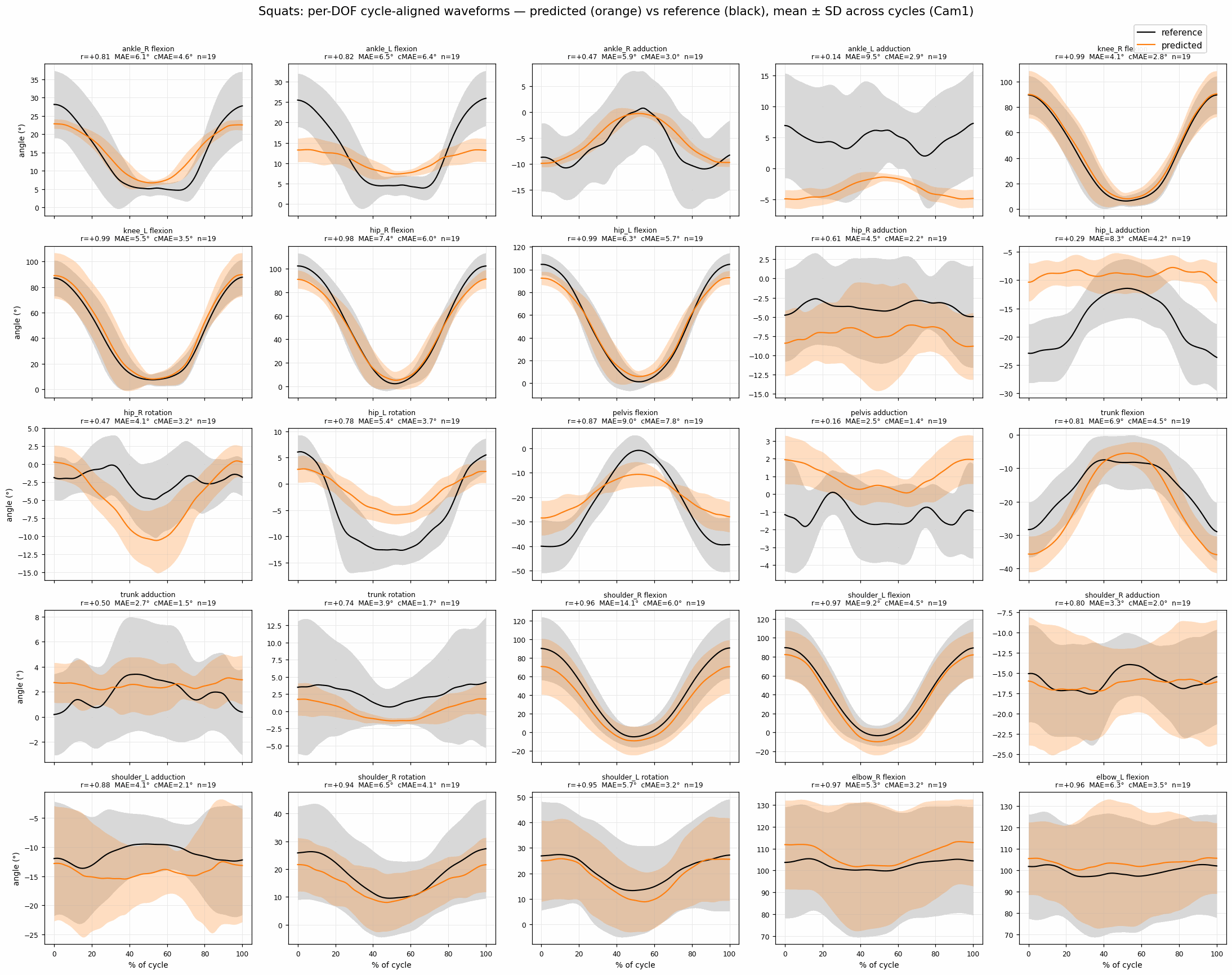}
  \caption{SAM 3D Body, squats: per-DOF joint-angle waveforms (degrees vs 0--100\% of cycle), predicted (orange) over reference (black), mean ±1 SD across cycles. Same layout as Figure B2.}
  \label{fig:sam3dbody_waveforms_squats}
\end{figure}

\begin{figure}[H]
  \centering
  \includegraphics[width=\textwidth]{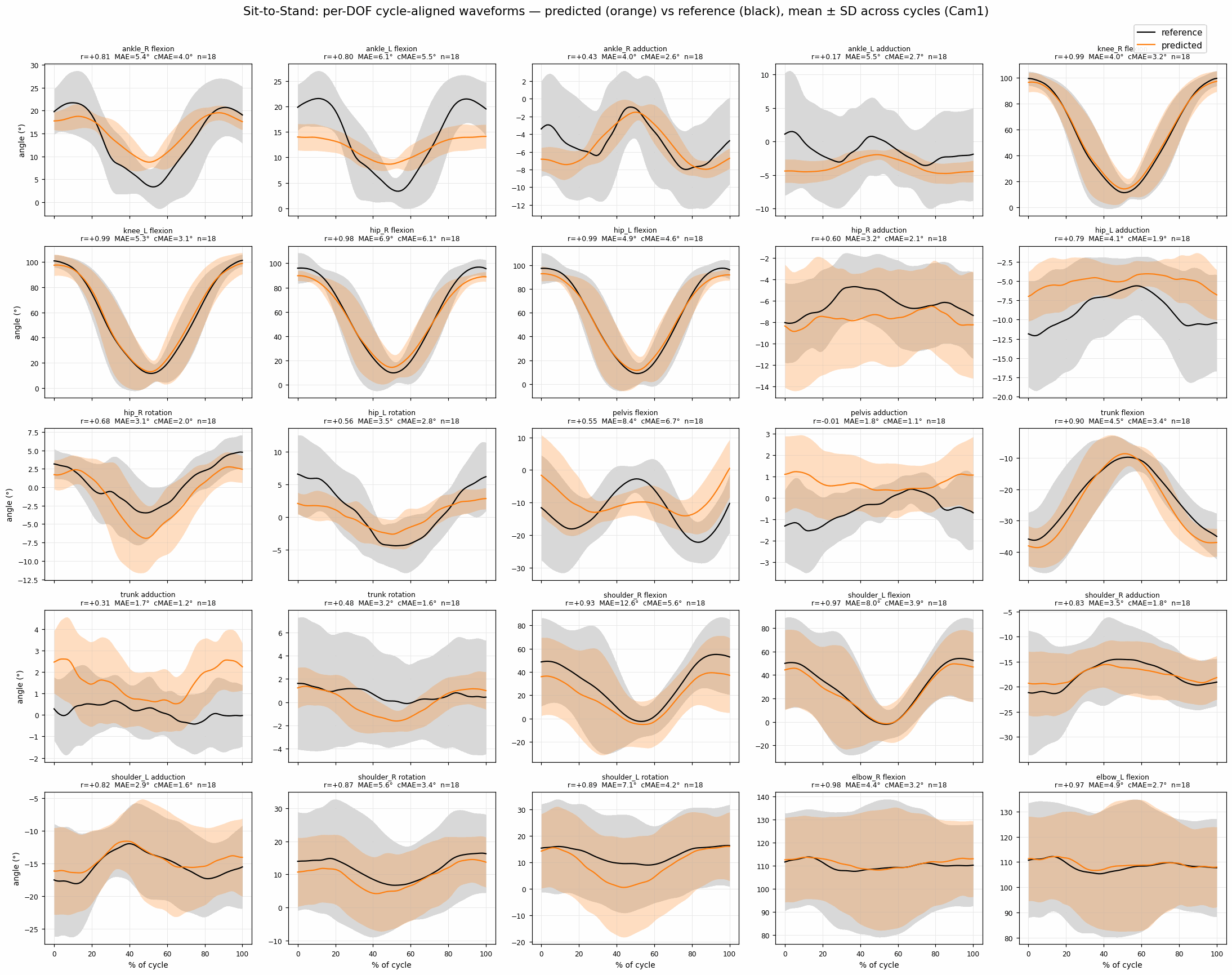}
  \caption{SAM 3D Body, sit-to-stand: per-DOF joint-angle waveforms (degrees vs 0--100\% of cycle), predicted (orange) over reference (black), mean ±1 SD across cycles. Same layout as Figure B3.}
  \label{fig:sam3dbody_waveforms_sit_to_stand}
\end{figure}

\begin{figure}[H]
  \centering
  \includegraphics[width=\textwidth]{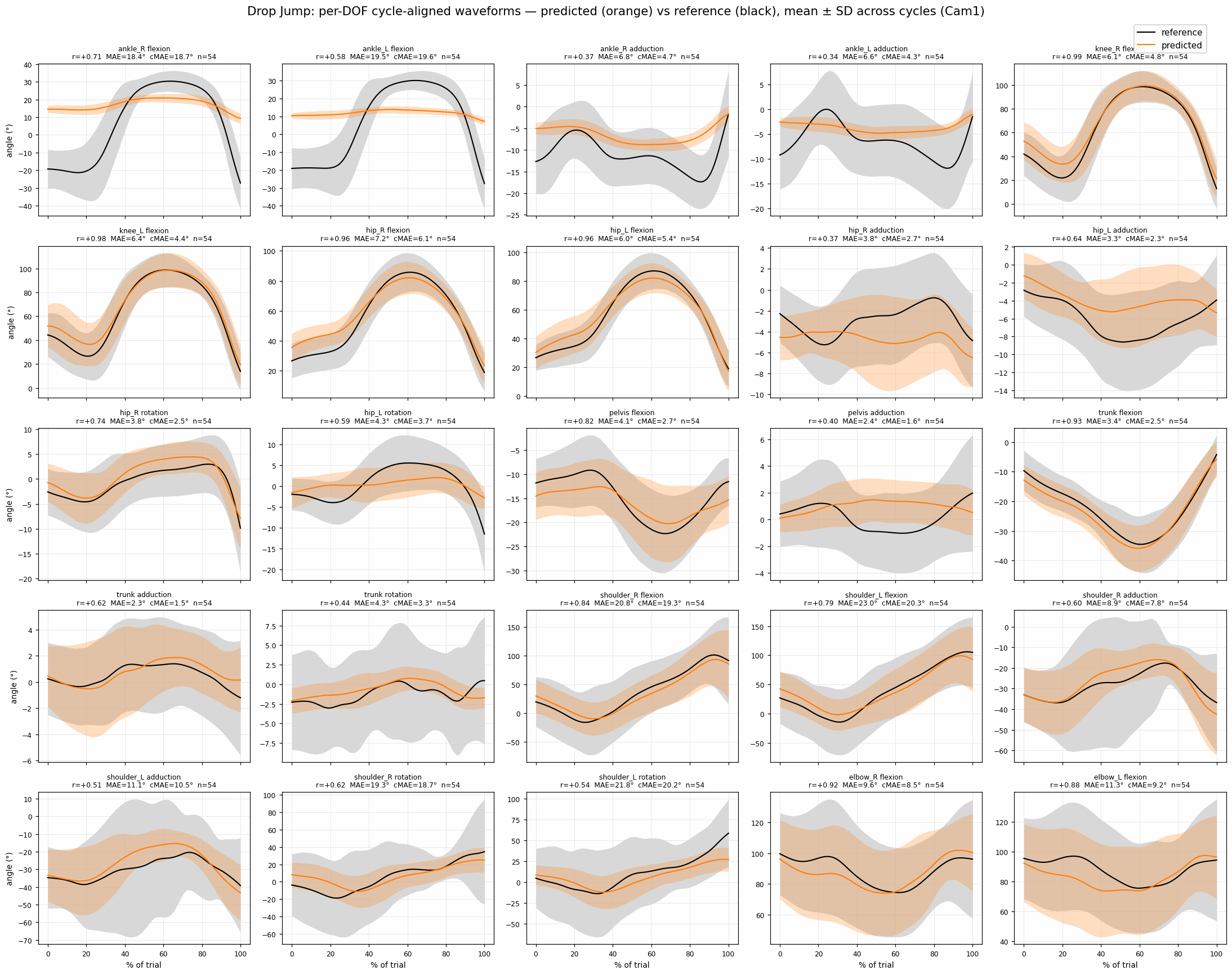}
  \caption{SAM 3D Body, drop jumps: per-DOF joint-angle waveforms (degrees vs 0--100\% of trial duration), predicted (orange) over reference (black), mean ±1 SD. Outside the validated envelope of Section 3.1, shown for completeness. Same layout as Figure B4.}
  \label{fig:sam3dbody_waveforms_drop_jump}
\end{figure}

\section{Joint definitions}

Each clinical joint is a parent-child segment pair in the body model's kinematic chain (Section 2.2). Table D1 lists the pairings; the pelvis is the root and has no parent.

\begin{longtable}{@{}lll@{}}
\caption{\textbf{Table D1.} Parent and child segment for each joint.}
\label{tab:d1}\\
\toprule\noalign{}
Joint & Parent segment & Child segment \\
\midrule\noalign{}
\endfirsthead
\toprule\noalign{}
Joint & Parent segment & Child segment \\
\midrule\noalign{}
\endhead
\bottomrule\noalign{}
\endlastfoot
hip & pelvis & upper leg \\
knee & upper leg & lower leg \\
shoulder & chest & upper arm \\
elbow & upper arm & forearm \\
ankle & lower leg & foot \\
trunk & pelvis & chest \\
pelvis & root & none \\
\end{longtable}

\section*{Author Contributions}

JK conceived the method, implemented the extraction pipeline and calibration procedure, performed the validation analysis, and wrote the manuscript. DH contributed to pipeline integration and data processing. JJ provided clinical context, contributed to interpretation of results, and reviewed the manuscript.

\section*{Competing Interests}

JK and DH are affiliated with Babon Innovations B.V., which has filed a patent application (NL4001444) covering the method described in this paper. JJ has no competing interests.

\section*{Acknowledgments}

We thank Uhlrich and colleagues for making the OpenCap LabValidation dataset publicly available \cite{uhlrich2023opencap}, and the Instituut Bewegingsstudies at HU University of Applied Sciences Utrecht for clinical input and partnership. This work stands on open-source research and tools, and we are grateful to their authors: OpenSim \cite{seth2018opensim} for the reference musculoskeletal modelling; the GEM-X estimator \cite{li2025genmo} and the SOMA skeleton convention \cite{saito2026soma} used in the main validation; and, for the cross-body-model validation in Appendix C, SAM 3D Body \cite{yang2026sam3d}, the MHR body model \cite{ferguson2025mhr}, the DINOv3 backbone \cite{simeoni2025dinov3}, and the open-source real-time C++/ONNX engine SAM3DBody-cpp \cite{qammaz2026sam3dbodycpp}.

\section*{Data Availability}

The validation dataset (OpenCap LabValidation) is publicly available at https://simtk.org/opencap. The calibration table is published in Appendix A.

\end{document}